\documentclass{article}

\usepackage{microtype}
\usepackage{graphicx,graphbox}
\usepackage{subcaption}
\usepackage{booktabs} % for professional tables
\usepackage{amssymb,mathtools}
\usepackage{todonotes}
\usepackage[nolist,nohyperlinks]{acronym}

\usepackage[utf8]{inputenc}

\usepackage{hyperref}

\usepackage[capitalise,nameinlink]{cleveref} % for easier referencing

\usepackage[accepted]{icml2021}

\newcommand{\E}[1]{\mathbb{E}\left[#1\right]}
\newcommand{\Esq}[1]{\mathbb{E}^2\left[#1\right]}
\newcommand{\Var}[1]{\operatorname{Var}\left[#1\right]}
\newcommand{\Ber}[1]{\operatorname{Ber}(#1)}
\newcommand{\Bin}[2]{\operatorname{B}(#1, #2)}
\newcommand{\matr}[1]{\mathbf{#1}}

\newcommand{\figvspace}{\vspace{-0.5em}}

\def\ie{\emph{i.e.,}}
\def\eg{\emph{e.g.,}}

\newacro{bnn}[BNN]{Bayesian neural network}
\acrodef{mse}[MSE]{mean squared error}
\acrodef{mcd}[MCD]{Monte-Carlo dropout}

\begin{document}

\twocolumn[
\icmltitle{Notes on the Behavior of MC Dropout}

% It is OKAY to include author information, even for blind
% submissions: the style file will automatically remove it for you
% unless you've provided the [accepted] option to the icml2021
% package.

% List of affiliations: The first argument should be a (short)
% identifier you will use later to specify author affiliations
% Academic affiliations should list Department, University, City, Region, Country
% Industry affiliations should list Company, City, Region, Country

% You can specify symbols, otherwise they are numbered in order.
% Ideally, you should not use this facility. Affiliations will be numbered
% in order of appearance and this is the preferred way.
%\icmlsetsymbol{equal}{*}

\begin{icmlauthorlist}
\icmlauthor{Francesco Verdoja}{aalto}
\icmlauthor{Ville Kyrki}{aalto}
\end{icmlauthorlist}

\icmlaffiliation{aalto}{School of Electrical Engineering, Aalto University,
Maarintie 8, Espoo, Finland}

\icmlcorrespondingauthor{Francesco Verdoja}{francesco.verdoja@aalto.fi}

% You may provide any keywords that you
% find helpful for describing your paper; these are used to populate
% the "keywords" metadata in the PDF but will not be shown in the document
\icmlkeywords{Machine Learning, ICML}

\vskip 0.3in
]

% this must go after the closing bracket ] following \twocolumn[ ...

% This command actually creates the footnote in the first column
% listing the affiliations and the copyright notice.
% The command takes one argument, which is text to display at the start of the footnote.
% The \icmlEqualContribution command is standard text for equal contribution.
% Remove it (just {}) if you do not need this facility.
%\printAffiliationsAndNotice{\icmlEqualContribution} % otherwise use the standard text.
\printAffiliationsAndNotice{This work was supported by the Strategic Research 
Council at Academy of Finland, decision 314180.

}

% Abstracts must be a single paragraph, ideally between 4--6 sentences long.
% Gross violations will trigger corrections at the camera-ready phase.
\begin{abstract}
    Among the various options to estimate uncertainty in deep neural networks,
    Monte-Carlo dropout is widely popular for its simplicity and effectiveness.
    However the quality of the uncertainty estimated through this method varies
    and choices in architecture design and in training procedures have to be
    carefully considered and tested to obtain satisfactory results. In this
    paper we present study offering a different point of view on the behavior
    of Monte-Carlo dropout, which enable us to observe a few interesting
    properties of the technique to keep in mind when considering its use for
    uncertainty estimation.
\end{abstract}

%%%%%%%%%%%%%%%%%%%%%%%%%%%%%%%%%%%%%%%%%%%%%%%%%%%%%%%%%%%%%%%%%%%%%%%%%%%%%%%%

\section{Introduction}
\label{sec:intro}

The increasing interest in the deployment of deep learning solutions in
real safety-critical applications ranging from hearthcare to robotics and
autonomous vehicles is making apparent the importance to properly estimate the
uncertainty of the predictions made by deep neural networks
\cite{kendallBayesianSegnetModel2017,loquercioGeneralFrameworkUncertainty2020}.

While most common neural network architectures only provide point estimates,
uncertainty can be evaluated with
\acp{bnn}~\cite{denkerTransformingNeuralnetOutput1990,
mackayPracticalBayesianFramework1992} where the deterministic weights used in
the majority of neural networks are replaced with distributions over the network
parameters. Although the formulation of \acp{bnn} is relatively easy in theory,
their use in practise for most complex problems is often unfeasible due to their
need to analytically evaluate the marginal probabilies during training which
becomes quickly intractable. Recently, variational inference methods have been
proposed as a practical alternative to \acp{bnn}, but most of these formulations
requires double the number of parameters of a network to represent its
uncertainty which leads to increased computational
costs~\cite{blundellWeightUncertaintyNeural2015,
galDropoutBayesianApproximation2016}.

Another very popular option to model uncertainty in deep neural networks is the
use of dropout as a way to approximate Bayesian variational
inference~\cite{galDropoutBayesianApproximation2016}. The simplicity of the key
idea of this formulation is one the main reasons for its popularity: by enabling
dropout not only in training but also during testing, and by doing several
forward passes through the network with the same input data, one can use the
distribution of the outputs of the different passes to estimate the first two
moments (mean and variance) of the predictive distribution. The mean is then
used as the estimate, and the variance as a measure of its uncertainty. This
technique, called \ac{mcd}, has proved effective to, \eg{} increase visual
relocalization accuracy~\cite{kendallModellingUncertaintyDeep2016} and semantic
segmentation performance on images~\cite{kendallBayesianSegnetModel2017}.
Despite its success and simplicity however, it has been recognized that the
quality of the uncertainty estimates is tied to parameter choices which need to
be calibrated to match the
uncertainty~\cite{osbandDeepExplorationBootstrapped2016,
pearceBayesianInferenceAnchored2018, bolukiLearnableBernoulliDropout2020,
caldeiraDeeplyUncertainComparing2020}. However, when using \ac{mcd} in practical
applications, architectural choices like where to insert the dropout layers, how
many to use, and the choice of dropout rate are often either empirically made
or set a priori \cite{kendallBayesianSegnetModel2017,
jungoUncertaintyAssistedBrainTumor2018, verdojaDeepNetworkUncertainty2019},
leading to possibly suboptimal performance.

In this work, we conduct a study providing some observations both in theory and
through experiments over the behavior of \ac{mcd}. The main contributions of
this work are a theoretical analysis over the behavior of \ac{mcd} on a simple
single-layer linear network, extending and correcting the discussion in
\cite{osbandRiskUncertaintyDeep2016}, and an experimental demonstration that the
properties found in theory apply to bigger non-linear networks as well. In the
discussion, we offer different intuitions over architecture design and training
choices for networks using \ac{mcd} for uncertainty estimation.

%%%%%%%%%%%%%%%%%%%%%%%%%%%%%%%%%%%%%%%%%%%%%%%%%%%%%%%%%%%%%%%%%%%%%%%%%%%%%%%%

\section{Behavior of MC Dropout}
\label{sec:theory}

In this section we expand and correct the intuitions first presented in
\cite{osbandRiskUncertaintyDeep2016} over the behavior of \ac{mcd}. In that
work, the author conducted a similar analysis but commented on the results only
partially. Here we correct some imprecisions in the way they computed the
optimal weights and analyse the results further. To this end, let us consider a
single-layer linear network
\begin{equation}\label{eq:linear}
    f = \sum_{k=1}^{K} d_k w_k
\end{equation}
with weights $w_k \in \mathbb{R}$ and dropouts $d_k \sim \Ber{p}$, \ie{} a
Bernoulli distribution with success probability $p$.  To note that here $p$
refers to $P(d_k = 1)$, while most dropout implementations require a dropout
probability parameter $p_{d} = P(d_k = 0) = 1 - p$.

Assuming all weights to converge to the same value $w$, which is to be expected
when using dropout, then
\begin{equation}\label{eq:varEf}
\begin{split}
    \E{f} &= \E{w \sum_{k=1}^{K} d_k} = w \E{\Bin{K}{p}} =\\
          &= w K p\\
    \Var{f} &= \Var{w \sum_{k=1}^{K} d_k} = w^2 \Var{\Bin{K}{p}} =\\
            &= w^2 K p (1-p)
\end{split}
\end{equation}
where $\Bin{K}{p}$ is a binomial distribution with $K$ trials and success
probability $p$.

Following \cite{osbandRiskUncertaintyDeep2016}, given a ground-truth $\{y_1,
\dots, y_n\}$ with average $\bar{y} \coloneqq \sum_{i=1}^{n} y_i/n $, minimizing
the \ac{mse} from the ground-truth means finding the minimum of 
\begin{equation}\label{eq:msesingle}
\begin{split}
    \E{(f - \bar{y})^2} &= \E{f^2-2f\bar{y}+\bar{y}^2}\\
    &= \E{f^2}-2\bar{y}\E{f}+\bar{y}^2\\
    &= \Var{f}+\Esq{f}-2\bar{y}\E{f}+\bar{y}^2\\
    %&= \Var{f}+(\bar{y}-\E{f})^2\\
    &= w^2 K p (1-p) + w^2 K^2 p^2 - 2\bar{y}wKp + \bar{y}^2\\
    &= w^2 K p (K p - p + 1) - 2\bar{y}wKp + \bar{y}^2
\end{split}
\end{equation}

This can be achieved by solving
\begin{equation}\label{eq:msederiv}
\begin{split}
    \frac{d}{dw}\E{(f - \bar{y})^2} &= 0\\
    2 w  K p (K p - p + 1) - 2\bar{y}Kp &= 0\\
    w (K p - p + 1) - \bar{y} &= 0
\end{split}
\end{equation}
which means the optimal weight is
\begin{equation}\label{eq:1w}
	w = \frac{\bar{y}}{Kp - p + 1} = \frac{\bar{y}}{K(1 - p_d) + p_d}
\end{equation}

Consequently, combining \cref{eq:varEf} and \cref{eq:1w}, we find that at
convergence:
\begin{equation}\label{eq:1ev}
\begin{split}
    \E{f} &= \frac{K p \bar{y}}{Kp - p + 1} 
           = \frac{K (1 - p_d) \bar{y}}{K(1 - p_d) + p_d}\\
    \Var{f} &= \frac{K p (1-p) \bar{y}^2}{(Kp - p + 1)^2} 
             = \frac{K p_d (1 - p_d) \bar{y}^2}{(K - K p_d + p_d)^2}
\end{split}
\end{equation}

From \cref{eq:1ev}, a few observation can be drawn: 
\begin{enumerate}
    \item while one would expect $\E{f} = \bar{y}$, the expected output of the
    network is actually introducing a bias. For big network however, this bias
    is negligible, since $Kp \approx Kp - p + 1$;
    \item the size of the variance of the posterior distribution generated by
    \ac{mcd} on this simple network depends on the interaction between the
    dropout rate $p_d$ and the model size $K$;
    \item the posterior distribution has no dependence on the amount of data
    $n$, nor the observed variance in the data, which means that it does not
    concentrate as more data is gathered;
    \item finally, the variance of the posterior distribution is proportional to
    $\bar{y}^2$, \ie{} the bigger the value to be estimated, the bigger the
    estimated model uncertainty.
\end{enumerate}

While we demonstrate here in theory that these properties of \ac{mcd} exist on
a very simple network, proving them following a similar thoeretical process on
bigger networks becomes quickly unfeasible.

For this reason, to try to understand how these interaction work on bigger more
realistic networks, we run different experiments, which we will present in the
following section.

\begin{table}
    \centering
    \caption{\label{tab:single} Comparison of single-layer network experiment 
    results versus the thoeretical expectation from \cref{eq:1w,eq:1ev}}
    \begin{tabular}{c c c c c c}
        \toprule
        & & \multicolumn{2}{c}{Thoeretical} & \multicolumn{2}{c}{Experimental}\\
        $p_d$ & Dataset & $w$ & $\Var{f}$ & $w$ & $\Var{f}$\\ 
        \midrule
        0.2 & $\matr{Y}'$  & 0.025 & 0.050 & 0.025 & 0.058\\
        0.2 & $\matr{Y}''$ & 0.025 & 0.050 & 0.026 & 0.076\\
        0.5 & $\matr{Y}'$  & 0.040 & 0.199 & 0.040 & 0.214\\
        0.5 & $\matr{Y}''$ & 0.040 & 0.199 & 0.040 & 0.249\\
        \bottomrule
    \end{tabular}
\end{table}

%%%%%%%%%%%%%%%%%%%%%%%%%%%%%%%%%%%%%%%%%%%%%%%%%%%%%%%%%%%%%%%%%%%%%%%%%%%%%%%%

\section{Experiments}
\label{sec:exp}

To verify that the properties found in \cref{sec:theory} hold true even for
bigger non-linear networks, we created the suite of experiments we report in
this section. These experiments have been implemented in \emph{pytorch
v0.3.0}\footnote{Source code for the experiments presented here is available at:
\href{https://github.com/aalto-intelligent-robotics/mc-dropout-notebooks}{github.com/aalto-intelligent-robotics/mc-dropout-notebooks}}.

\subsection{Single-layer linear network}
\label{sec:exp1}

In this first experiment we want to empirically verify the theoretical behavior
observed in \cref{sec:theory}. In particualar, we implemented two versions of
the single-layer linear network in \cref{eq:linear} with $K = 500$, one with
$p_d = 0.2$ and one with $p_d = 0.5$ respectively. Once more, note that the
dropout probability $p_d = 1 - p$. We created two ground-truth datasets $
\matr{Y}' = \{y'_1, \dots, y'_n\}$ and $ \matr{Y}'' = \{y''_1, \dots, y''_n\}$
with $n=3200$, $y'_i \sim \mathcal{N}(10, 1)$, and $y''_i \sim \mathcal{N}(10,
10)$. We trained both networks on both datasets for 600 epochs with \ac{mse}
loss and adam optimizer. As a side note, training with \ac{mse} holds the same
results as training with log-likelihood, since the minimum of the two losses is
in the same position in this case. After training, we run multiple forward
passes of the networks with dropout active to obtain one million samples to
empirically estimate $\Var{f}$.

\begin{figure}
    \centering
    \begin{subfigure}{.48\linewidth}
        \includegraphics[width=\linewidth]{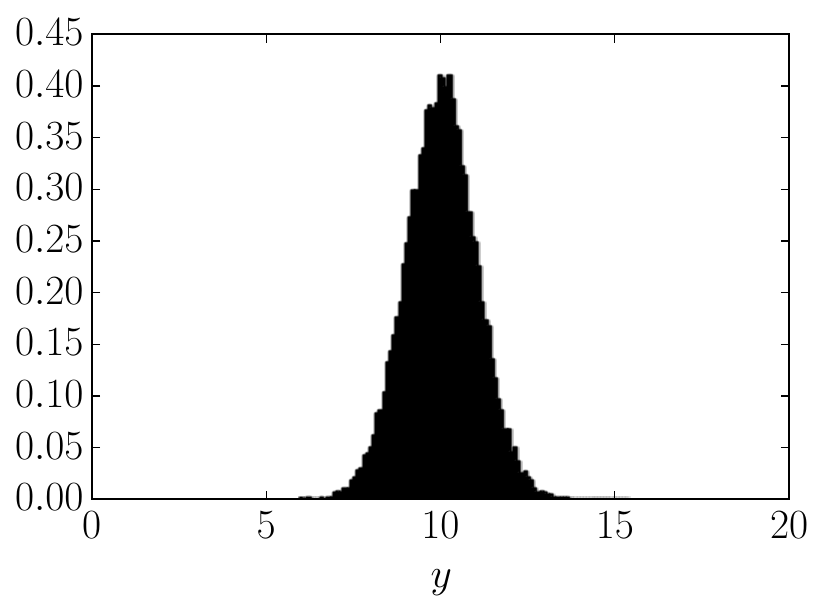}
        \caption{\label{fig:single_a}$\matr{Y}'$ ($\mathcal{N}(10,1)$)}
    \end{subfigure}
    \begin{subfigure}{.48\linewidth}
        \includegraphics[width=\linewidth]{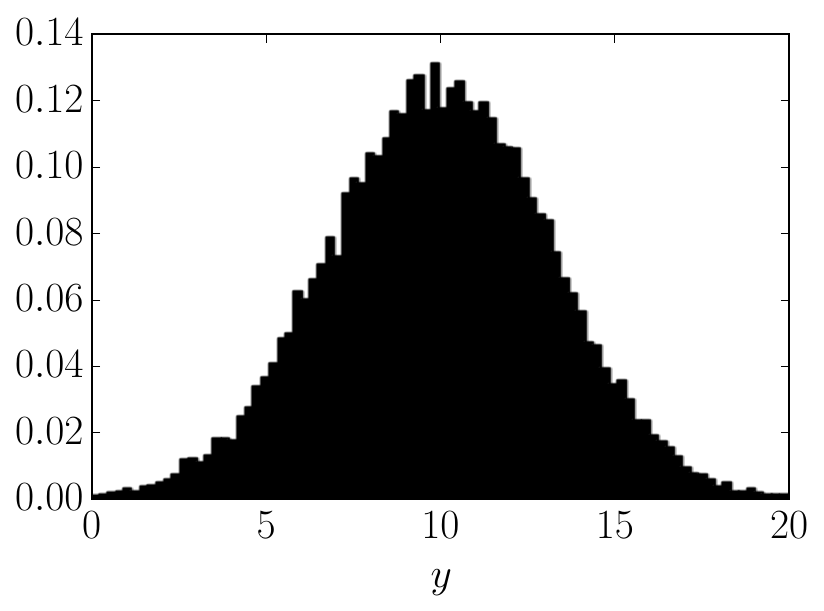}
        \caption{\label{fig:single_b}$\matr{Y}''$ ($\mathcal{N}(10,10)$)}
    \end{subfigure}
    \begin{subfigure}{.48\linewidth}
        \includegraphics[width=\linewidth]{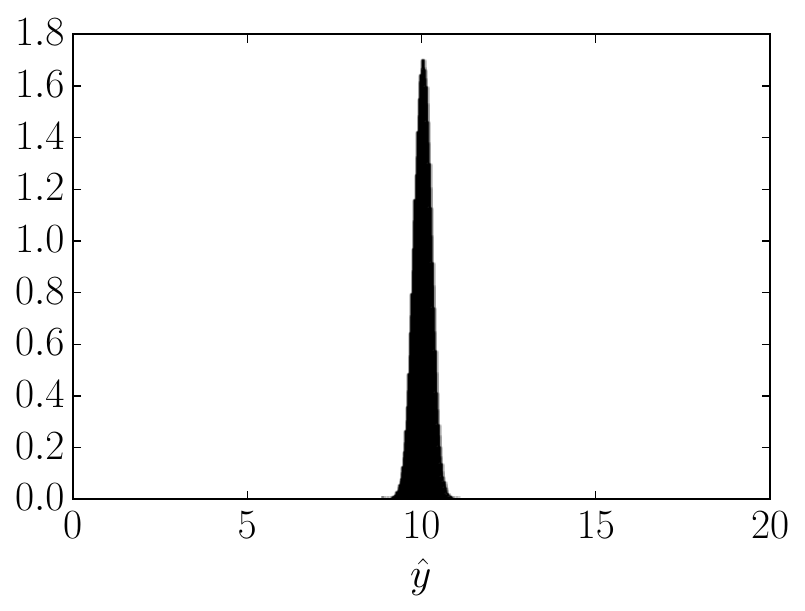}
        \caption{\label{fig:single_c}$p_d = 0.2$, on $\matr{Y}'$}
    \end{subfigure}
    \begin{subfigure}{.48\linewidth}
        \includegraphics[width=\linewidth]{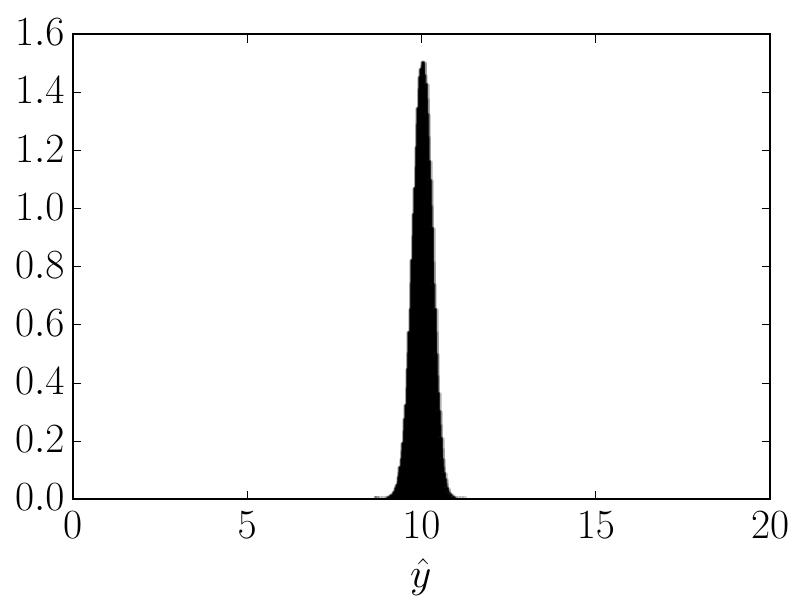}
        \caption{\label{fig:single_d}$p_d = 0.2$, on $\matr{Y}''$}
    \end{subfigure}
    \begin{subfigure}{.48\linewidth}
        \includegraphics[width=\linewidth]{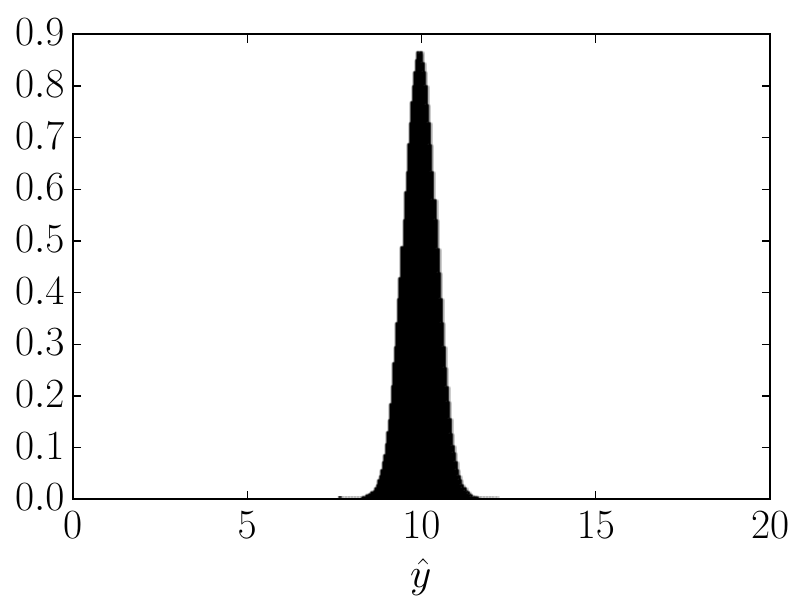}
        \caption{\label{fig:single_e}$p_d = 0.5$, on $\matr{Y}'$}
    \end{subfigure}
    \begin{subfigure}{.48\linewidth}
        \includegraphics[width=\linewidth]{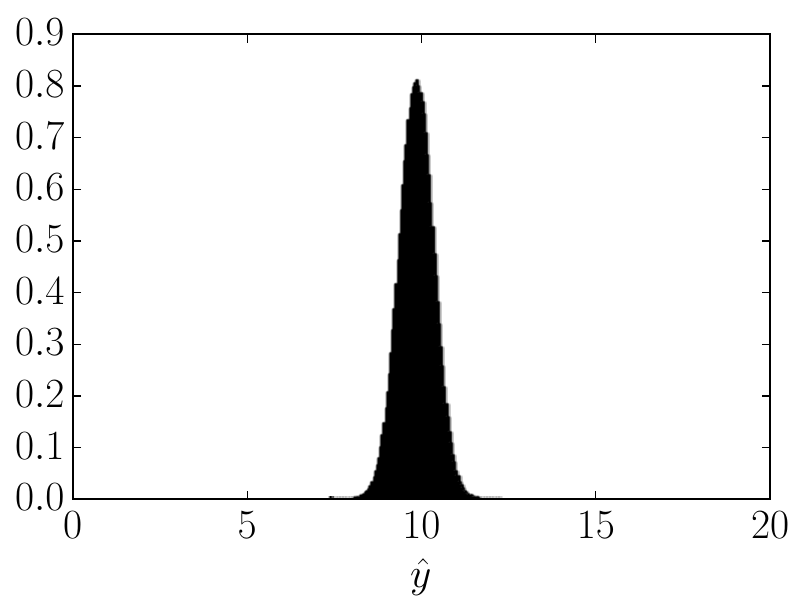}
        \caption{\label{fig:single_f}$p_d = 0.5$, on $\matr{Y}''$}
    \end{subfigure}
    \caption{Ground-truth distributions (a, b) and corresponding \ac{mcd} 
    output distributions for single-layer linear networks with $p_d = 0.2$ (c,
    d) and $p_d = 0.5$ (e, f) respectively.}
    \label{fig:single}
    \vspace{-1em}
\end{figure}

In \cref{tab:single}, we can see that the values for $w$ and $\Var{f}$ found
experimentally match the theoretical ones computed using \cref{eq:1w,eq:1ev},
aside for noise introduced by the sampling process. Moreover, they confirm how
changing the parameter $p_d$ afftects the model uncertainty, while changing
observed variance in the data (from $\sigma = 1$ in $\matr{Y}'$, to $\sigma =
10$ in $\matr{Y}''$) does not. This phenomenon is also clearly visible in
\cref{fig:single} where it can be noticed how the variance of the output
distribution of \ac{mcd} is impacted by $p_d$ (compare
\cref{fig:single_c,fig:single_e}, and \cref{fig:single_d,fig:single_f}) while
being unaffected by the variance in the dataset it trained on (compare
\cref{fig:single_c,fig:single_d}, and \cref{fig:single_e,fig:single_f}).

\subsection{Non-linear networks}
\label{sec:exp2}

\begin{figure}
    \centering
    \includegraphics[width=\linewidth]{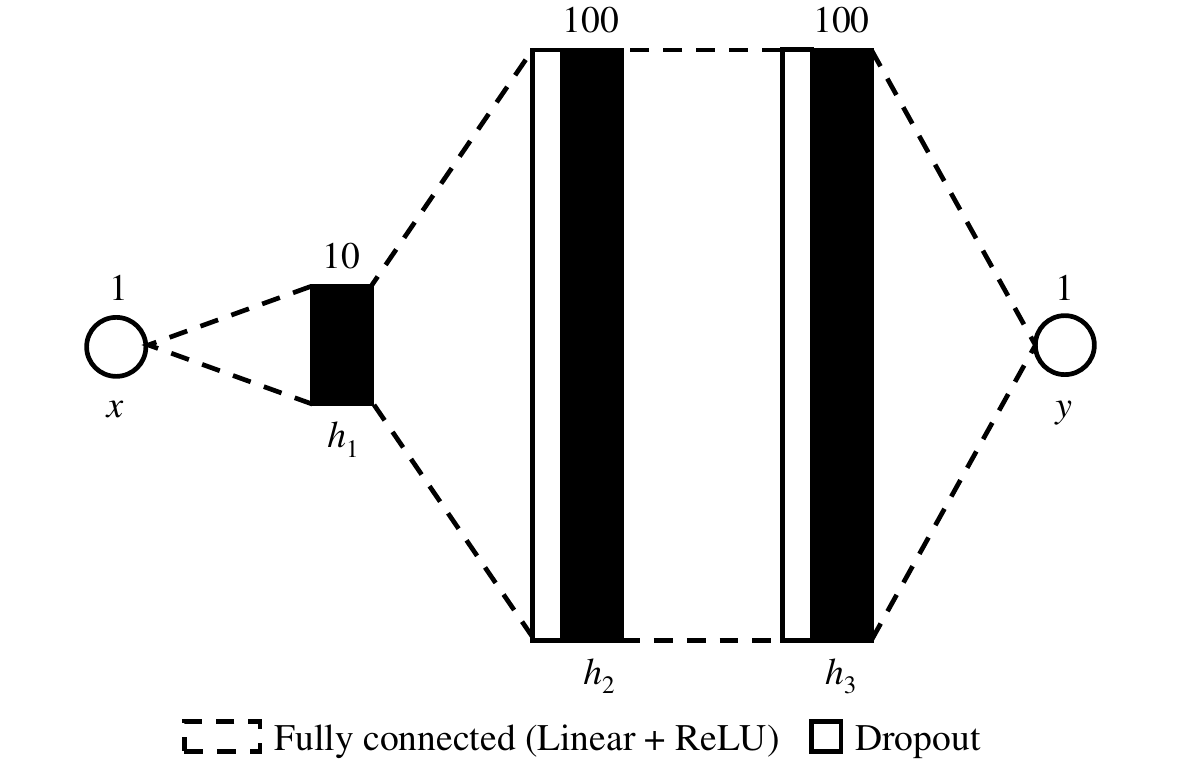}
    \caption{\label{fig:fcnet} Fully connected network used in the experiments.}
    \vspace{-1em}
\end{figure}

To verify that effects similar to those presented in the previous section can
still be observed after introducing more layers and non-linearities, we
performed a similar experiment on the fully connected neural network shown in
\cref{fig:fcnet}. While this network is still smaller then most real networks,
we decided for this size because it had enough complexity while still being
simple enough to analyze. In all the following experiments we trained for 1000
epochs with \ac{mse} loss and adam optimizer on a dataset of 32000 samples. We
trained two variants of the network with dropout rate $p_d = 0.2$ and $p_d =
0.5$ respectively. As it can be seen in the figure, for these experiments the
network has one input $x$ and produces one output $y$, effectively approximating
a function $f: \mathbb{R} \to \mathbb{R}$.

\begin{figure}
    \centering
    \begin{subfigure}{.48\linewidth}
        \includegraphics[width=\linewidth]{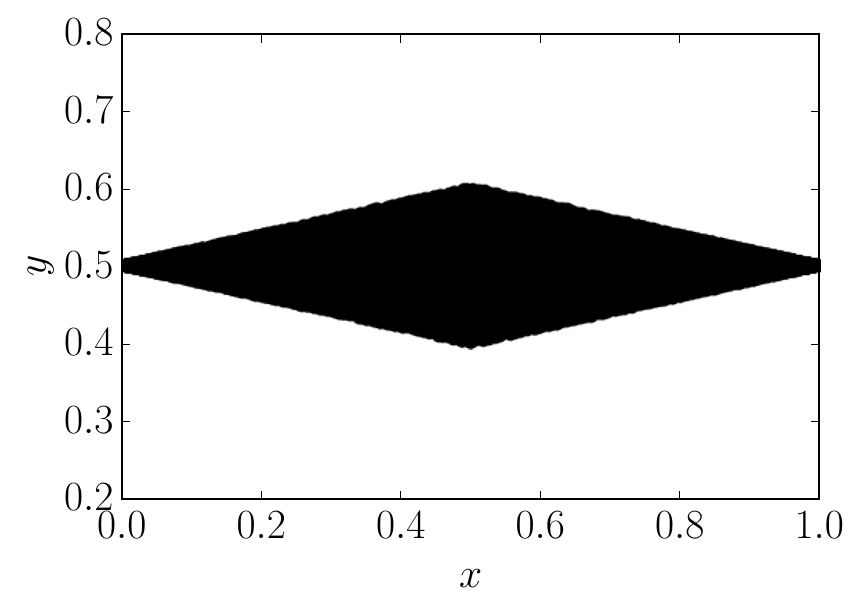}
        \caption{\label{fig:diamond_dist}Ground-truth function}
    \end{subfigure}
    \begin{subfigure}{.48\linewidth}
        \includegraphics[width=\linewidth]{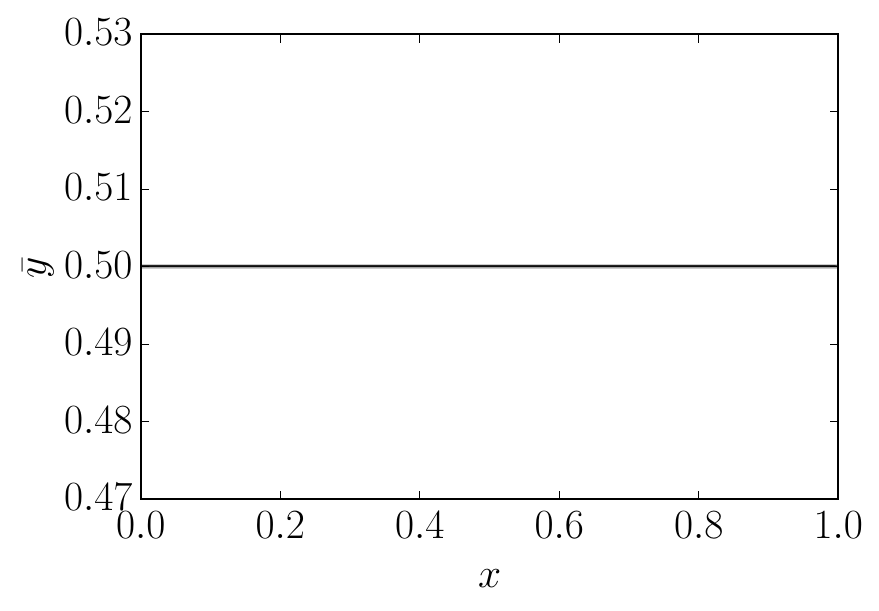}
        \caption{\label{fig:diamond_bias}with bias, $p_d = 0.5$}
    \end{subfigure}
    \begin{subfigure}{.48\linewidth}
        \includegraphics[width=\linewidth]{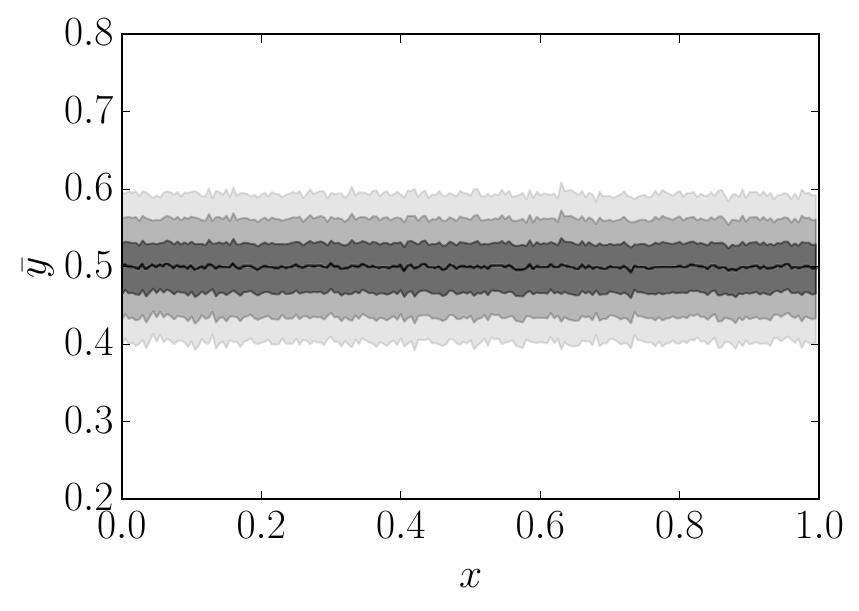}
        \caption{\label{fig:diamond_0.2}no bias, $p_d = 0.2$}
    \end{subfigure}
    \begin{subfigure}{.48\linewidth}
        \includegraphics[width=\linewidth]{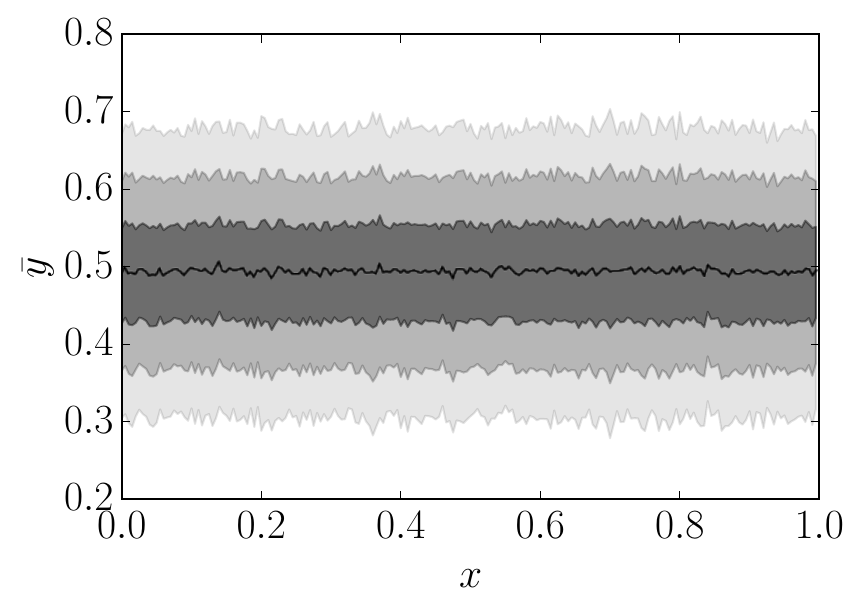}
        \caption{\label{fig:diamond_0.5}no bias, $p_d = 0.5$}
    \end{subfigure}
    \caption{\label{fig:diamond} Results obtained by training a non-linear
    network on the dataset shown in (a); different variants of the network have
    been trained: one with a bias term in the last linear layer (b) and two
    without bias in the last linear layer and with different dropout rates (c,
    d). In (a) each dot represents a datapoint; in (b, c, d) the line represents
    the average output of 300 forward passes through the network, with the
    shaded areas representing $\sigma$, $2\sigma$, and $3\sigma$ respectively.}
    \figvspace{}
\end{figure}

\cref{fig:diamond} shows the results produced by the network on a noisy constant
function where for each input $x \in [0,1]$ multiple possible outputs $y$ are
present in the dataset while $\forall x, \E{y} = 0.5$ (\cref{fig:diamond_dist}).
We first show the behavior of the network when the last linear layer has a bias
term in \cref{fig:diamond_bias}. It can be seen that the network is nullifing
the variance completely. It is indeed setting all weights to 0 and encoding the
desired (constant) output in the bias. If we remove the bias from the last
linear layer (\cref{fig:diamond_0.2,fig:diamond_0.5}) we obtain constant
variance, proportional to the dropout rate.

\begin{figure}
    \centering
    \begin{subfigure}{.48\linewidth}
        \includegraphics[width=\linewidth]{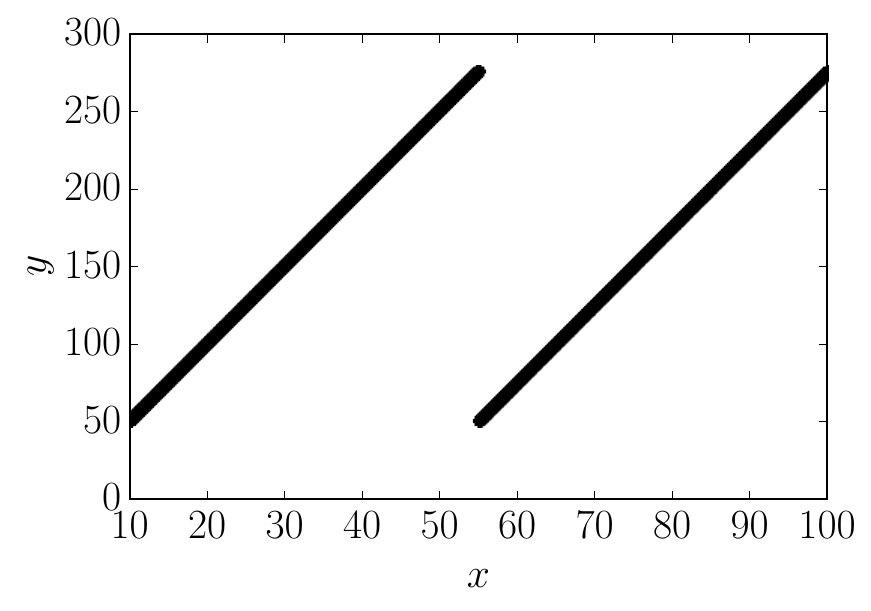}
        \caption{\label{fig:saw_dist}Ground-truth function}
    \end{subfigure}
    \begin{subfigure}{.48\linewidth}
        \includegraphics[width=\linewidth]{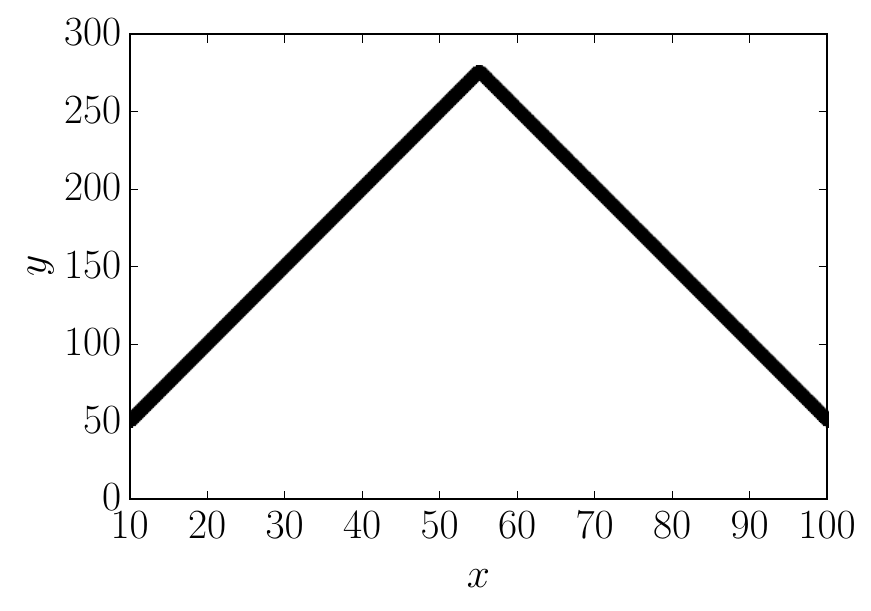}
        \caption{\label{fig:triangle_dist}Ground-truth function}
    \end{subfigure}
    \begin{subfigure}{.48\linewidth}
        \includegraphics[width=\linewidth]{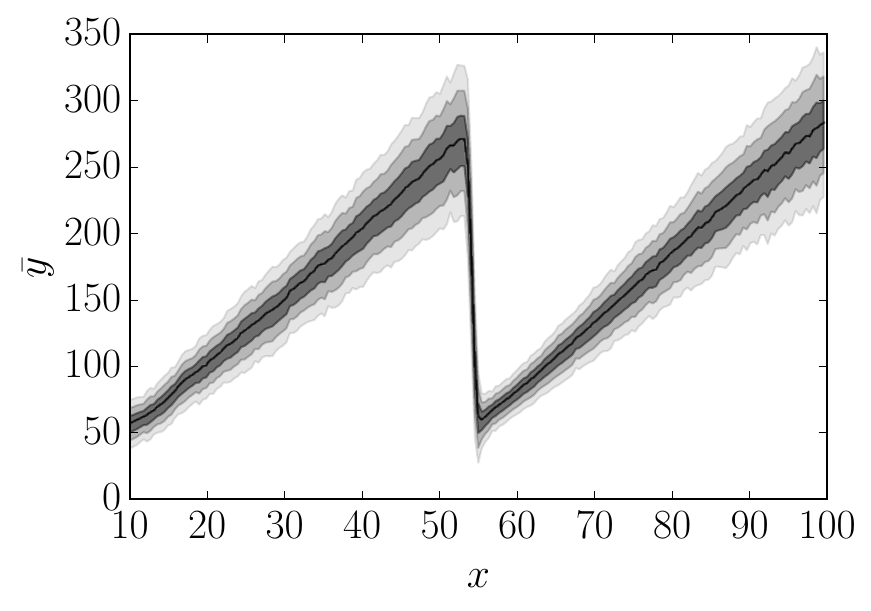}
        \caption{\label{fig:saw_0.2}$p_d = 0.2$}
    \end{subfigure}
    \begin{subfigure}{.48\linewidth}
        \includegraphics[width=\linewidth]{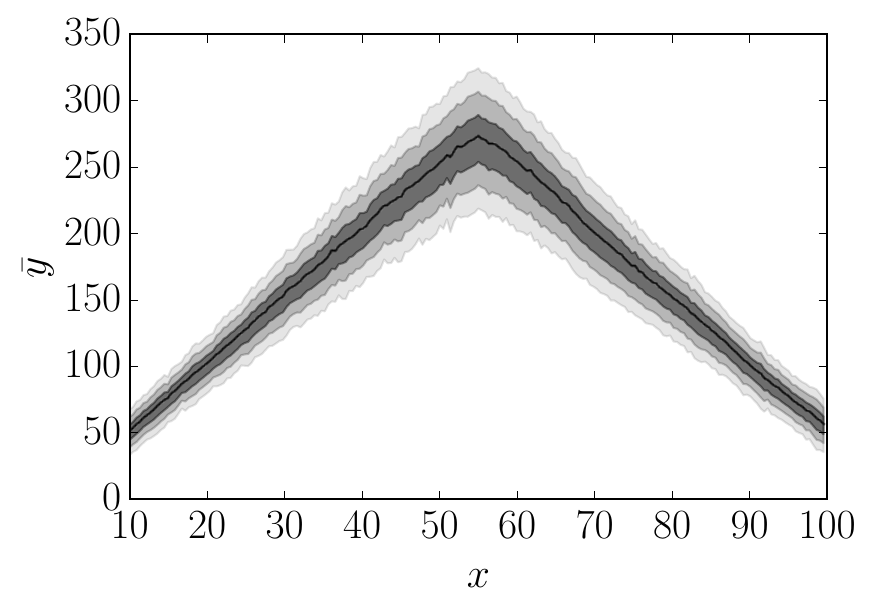}
        \caption{\label{fig:triangle_0.2}$p_d = 0.2$}
    \end{subfigure}
    \begin{subfigure}{.48\linewidth}
        \includegraphics[width=\linewidth]{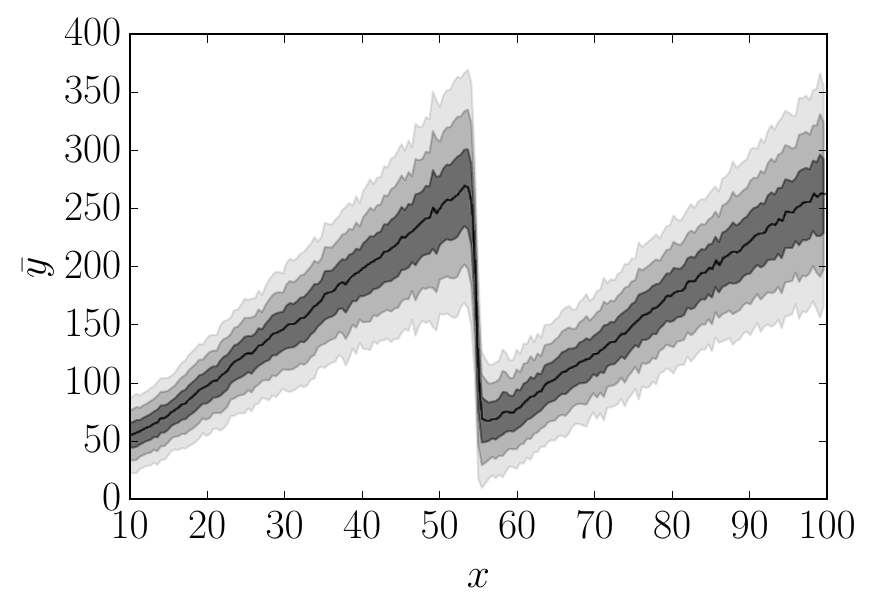}
        \caption{\label{fig:saw_0.5}$p_d = 0.5$}
    \end{subfigure}
    \begin{subfigure}{.48\linewidth}
        \includegraphics[width=\linewidth]{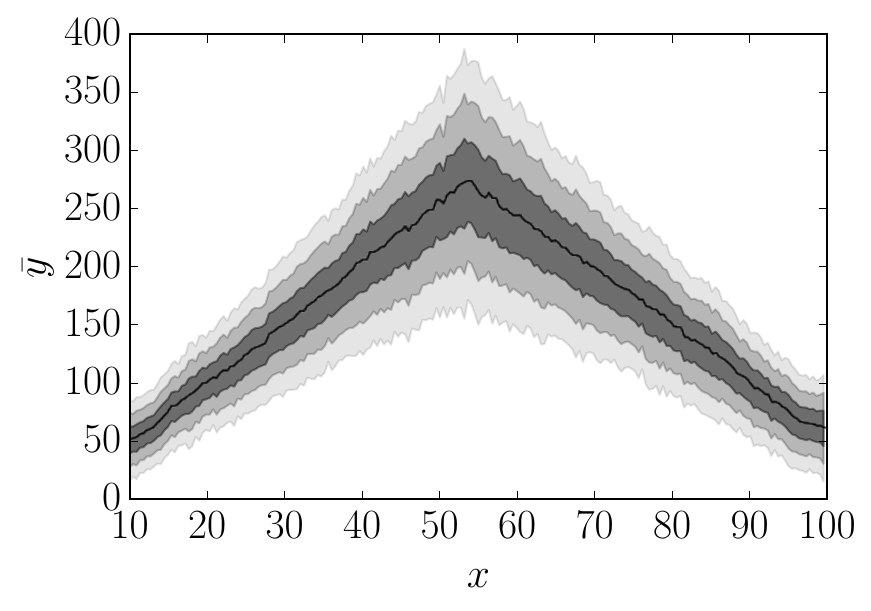}
        \caption{\label{fig:triangle_0.5}$p_d = 0.5$}
    \end{subfigure}
    \caption{\label{fig:variance} Results obtained by training a non-linear
    network on the dataset shown in (a, b); two variants of the network have
    been trained: with dropout rate $p_d = 0.2$ (c,d) and $p_d = 0.5$ (e, f). In
    (a, b) each dot represents a datapoint; in (c, d, e, f) the line represents
    the average output of 300 forward passes through the network, with the
    shaded areas representing $\sigma$, $2\sigma$, and $3\sigma$ respectively.}
    \figvspace{}
\end{figure}

\cref{fig:variance} shows another experiment, conducted on different functions.
In this case the datasets have no noise, \ie{} for each $x$, all samples in the
dataset have the same corresponding $y$. All the functions and network variants
we tested are available as additional material, however, due to space
limitations, in here we present only two functions
(\cref{fig:saw_dist,fig:triangle_dist}) and only the results for networks where
the bias has been disabled on the last linear layer. All results not reported
here lead to the same conclusions. From these results, the last of the
properties presented in \cref{sec:theory} can be noticed, namely the fact that
the size of the uncertainty scales proportionally to the value of $y$. These
also confirm one more time the dependence of the uncertainty from the dropout
parameter $p_d$.

%%%%%%%%%%%%%%%%%%%%%%%%%%%%%%%%%%%%%%%%%%%%%%%%%%%%%%%%%%%%%%%%%%%%%%%%%%%%%%%%

\vspace{-0.4em}
\section{Discussion}
\label{sec:disc}

The results presented in \cref{sec:exp} confirm that the behaviors of \ac{mcd}
that we theoretically observed on a single-layer linear network are still
present even as the size and complexity of the network grows. When looking at
these results, we are able to make a few observations.

First of all, all experiments provided an additional demonstration that the
choice of dropout rate $p_d$ is crucial and that in itself the epistemic
uncertainty estimate provided by \ac{mcd} is not affected by the amount of data
available during training or its variance, which is a good reminder that the
uncertainty estimate produced by \ac{mcd} is not calibrated and that the the
dropout rate has to be adjusted to match the epistemic uncertainty. In practise,
this is usually done through grid search, although several methods have been
proposed to learn the parameter during training, \eg{}
\cite{galConcreteDropout2017, phanCalibratingUncertaintiesObject2018,
bolukiLearnableBernoulliDropout2020}. 

Another interesting aspect that emerged from our study was the scaling of the
uncertainty based on the value of the network output. This effect, if not taken
into account, can lead to degraded quality of the uncertainty estimates, which
we empiracally noticed in the past in a practical robotic application
\cite{verdojaDeepNetworkUncertainty2019}. This effect is particularly visible
when dropout is applied before the last linear layer of the network, as is the
case in the the experiments presented here. If more non-linear layers are added
after the last dropout layer the non-linearities can reduce this effect. This
would suggest that in applications with widely varying outputs, it could be
useful to limit dropout only to the inner parts of the network, justifying what
other studies have found experimentally to helps improve performance
\cite{kendallBayesianSegnetModel2017, jungoUncertaintyAssistedBrainTumor2018}.

%%%%%%%%%%%%%%%%%%%%%%%%%%%%%%%%%%%%%%%%%%%%%%%%%%%%%%%%%%%%%%%%%%%%%%%%%%%%%%%%

\vspace{-0.4em}
\section{Conclusions}
\label{sec:conc}

In this paper we presented a different point of view on the behavior of
\ac{mcd} that enabled us to show both theoretically and experimentally different
properties of that approach, in particular the dependency of the uncertainty
estimates from the dropout rate and from the value the network is trying to
estimate. Our intent was to propose a framework to demonstrate these property in
a very understandable fashion, to help in both architecture design and training
procedures when using \ac{mcd}.

%%%%%%%%%%%%%%%%%%%%%%%%%%%%%%%%%%%%%%%%%%%%%%%%%%%%%%%%%%%%%%%%%%%%%%%%%%%%%%%%

\bibliography{refs}
\bibliographystyle{icml2021}

\newpage

%%%%%%%%%%%%%%%%%%%%%%%%%%%%%%%%%%%%%%%%%%%%%%%%%%%%%%%%%%%%%%%%%%%%%%%%%%%%%%%%

\begin{figure*}
    Additional results obtained by training a non-linear network on
    different datasets (row a). Different variants of the network have been
    trained: one with $p_d = 0.2$ and bias term in the last linear layer (rows
    b, c), one with $p_d = 0.2$ without bias term in the last linear layer (rows
    d, e), one with $p_d = 0.5$ and bias term in the last linear layer (rows f,
    g), and one with $p_d = 0.5$ without bias term in the last linear layer
    (rows h, i). In (rows a, d, h) each dot represents a datapoint; in (rows c,
    e, g, i) the line represents the average output of 300 forward passes
    through the network, with the shaded areas representing $\sigma$, $2\sigma$,
    and $3\sigma$ respectively.\\[.3em]    
    (a)%
    \includegraphics[align=c,width=.19\linewidth]{plots/diamond_0.2_nobias_1.pdf}%
    \includegraphics[align=c,width=.19\linewidth]{plots/saw_0.2_bias_1.pdf}%
    \includegraphics[align=c,width=.19\linewidth]{plots/triangle_0.2_bias_1.pdf}%
    \includegraphics[align=c,width=.19\linewidth]{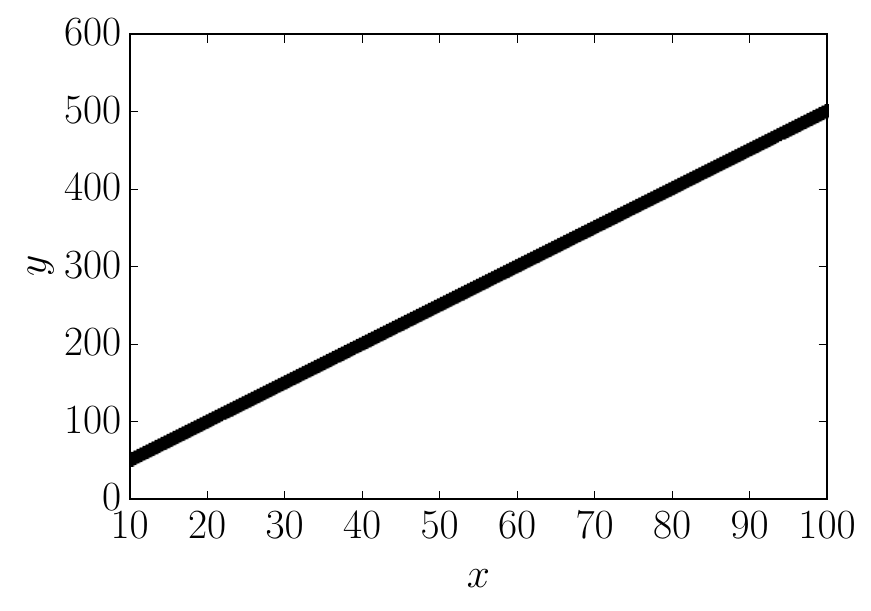}%
    \includegraphics[align=c,width=.19\linewidth]{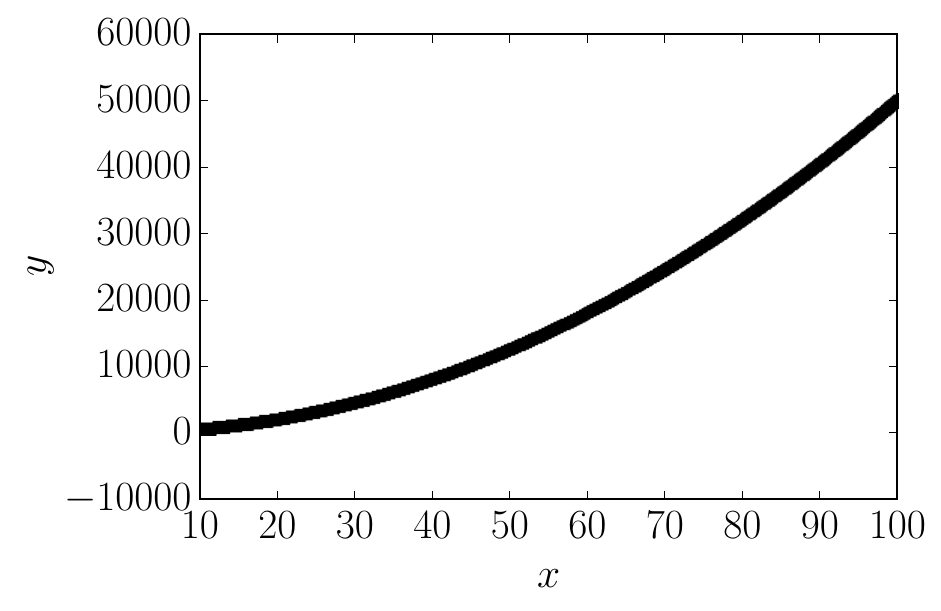}%
    \\
    (b)%
    \includegraphics[align=c,width=.19\linewidth]{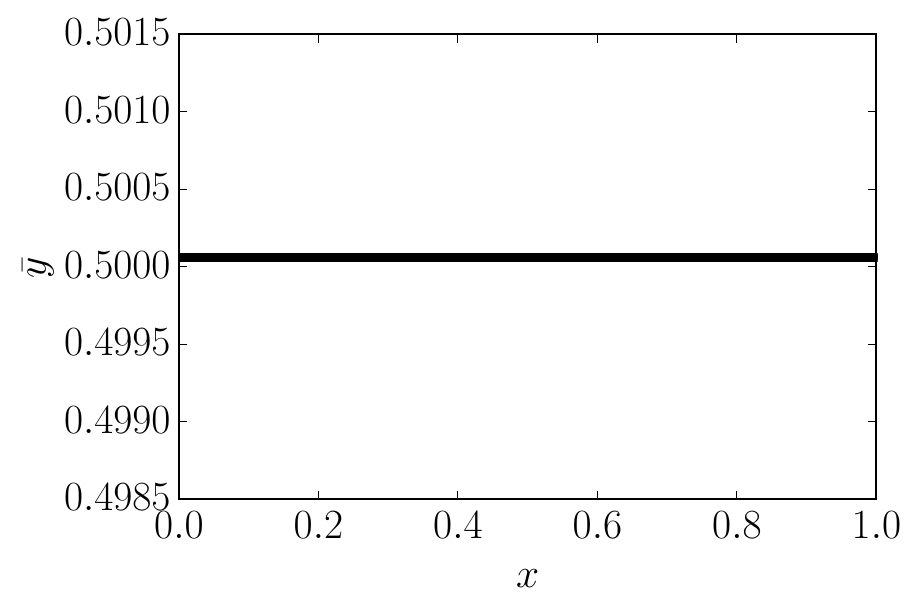}%
    \includegraphics[align=c,width=.19\linewidth]{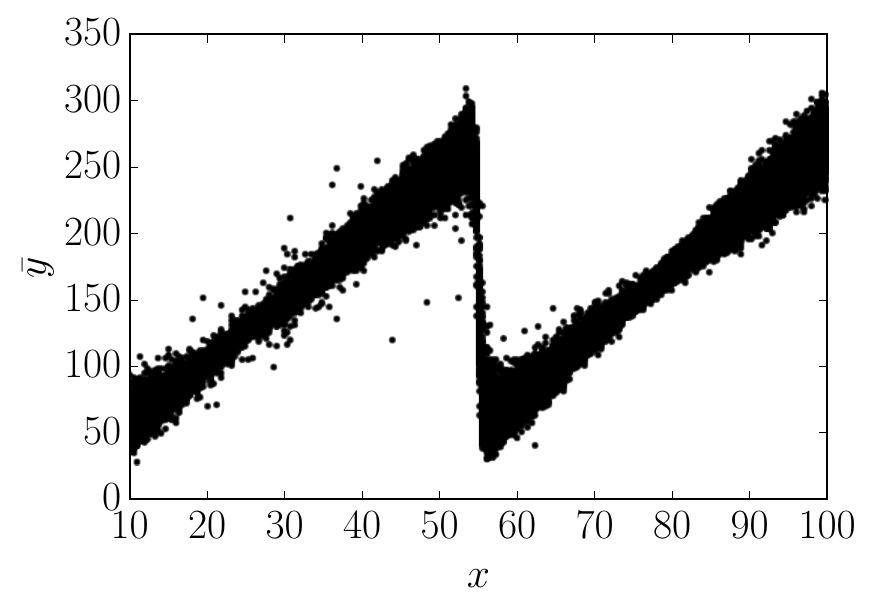}%
    \includegraphics[align=c,width=.19\linewidth]{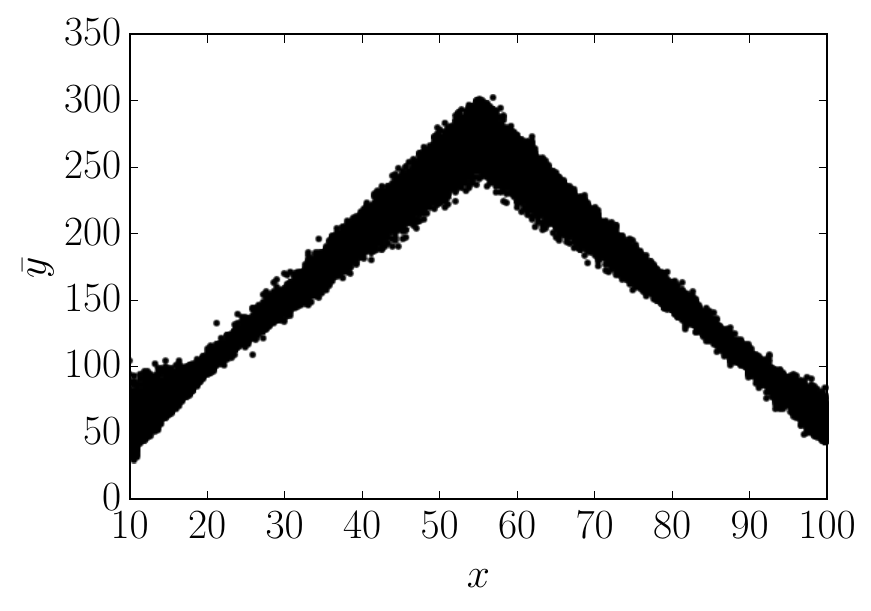}%
    \includegraphics[align=c,width=.19\linewidth]{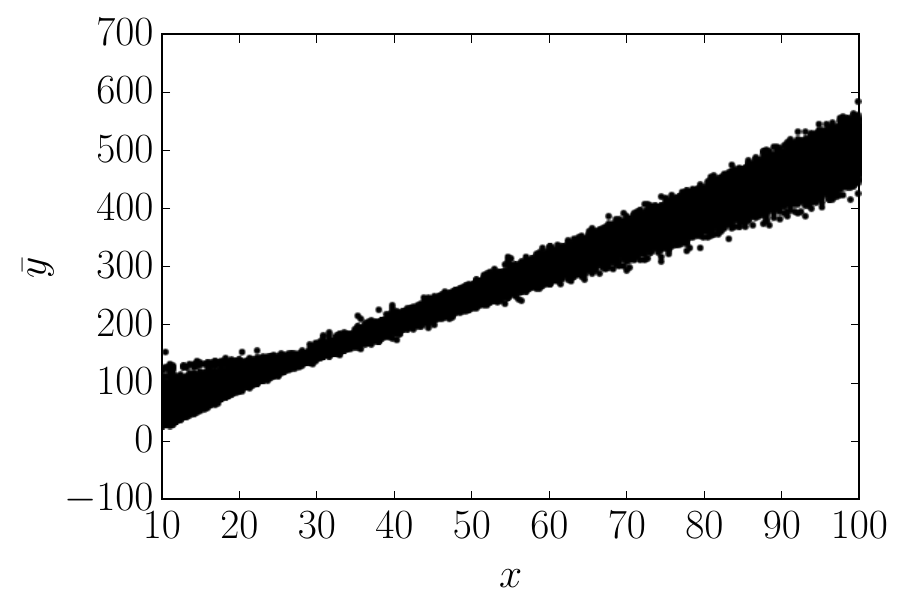}%
    \includegraphics[align=c,width=.19\linewidth]{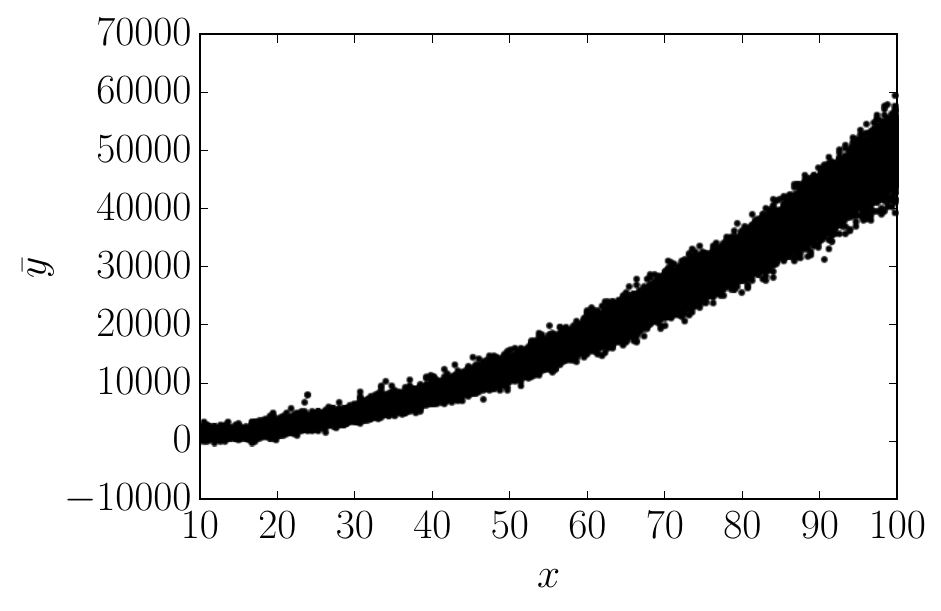}%
    \\
    (c)%
    \includegraphics[align=c,width=.19\linewidth]{plots/diamond_0.2_bias_3.pdf}%
    \includegraphics[align=c,width=.19\linewidth]{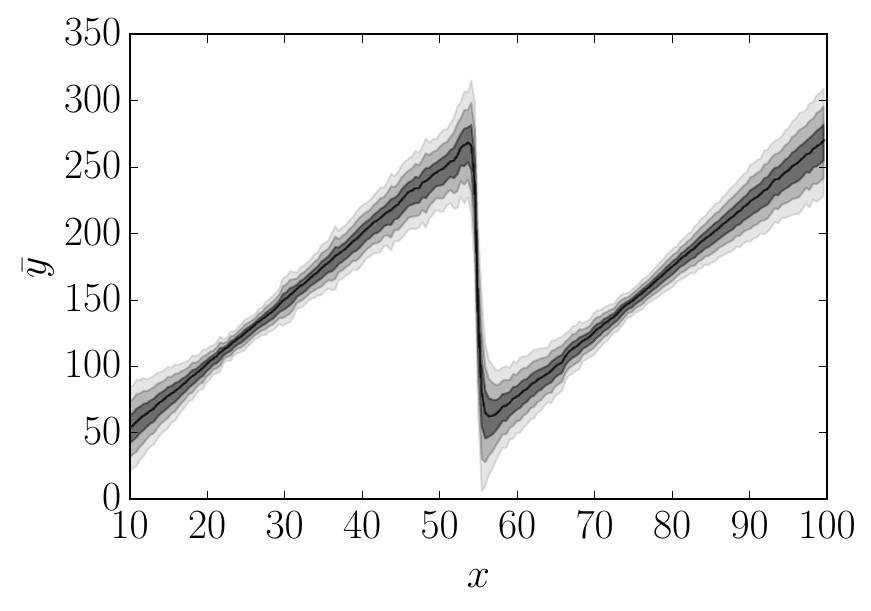}%
    \includegraphics[align=c,width=.19\linewidth]{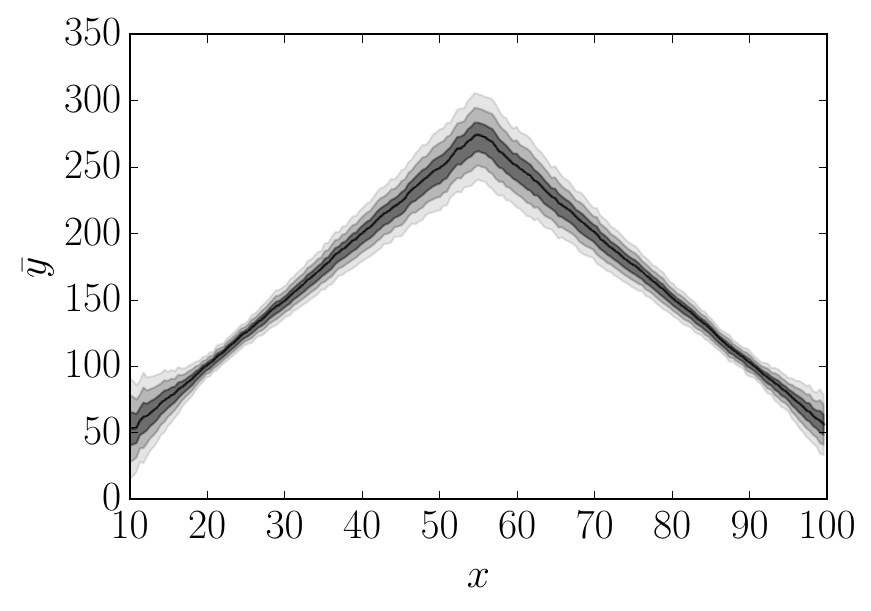}%
    \includegraphics[align=c,width=.19\linewidth]{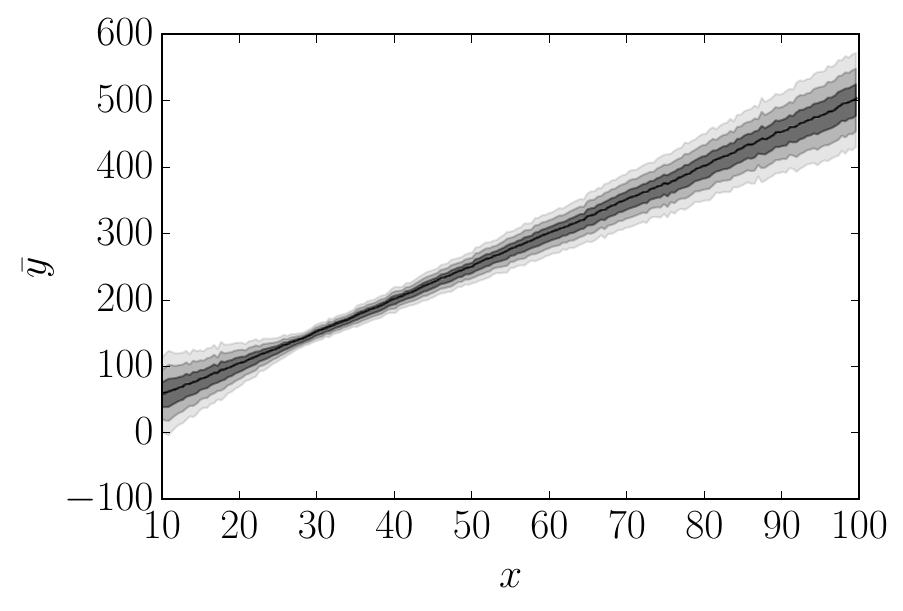}%
    \includegraphics[align=c,width=.19\linewidth]{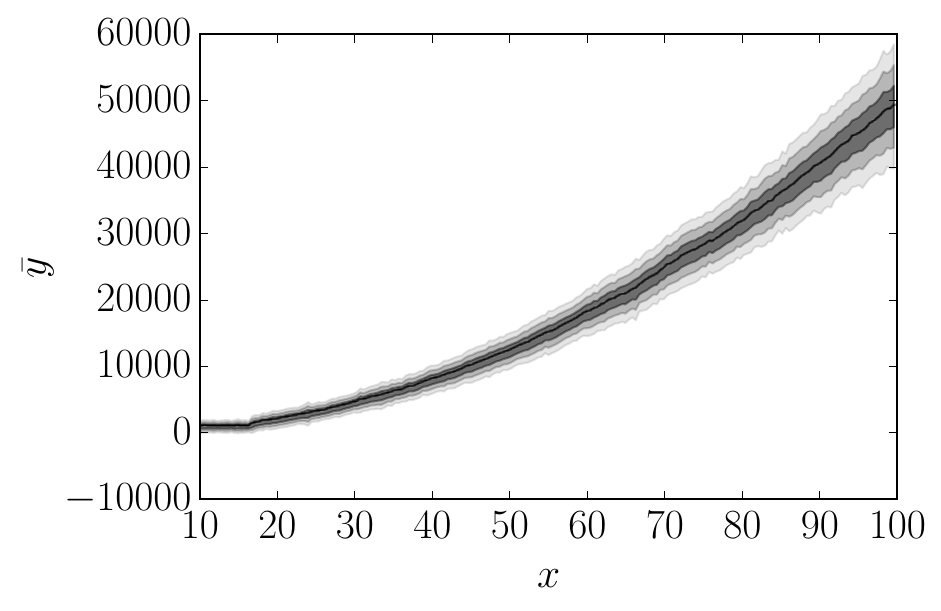}%
    \\
    (d)%
    \includegraphics[align=c,width=.19\linewidth]{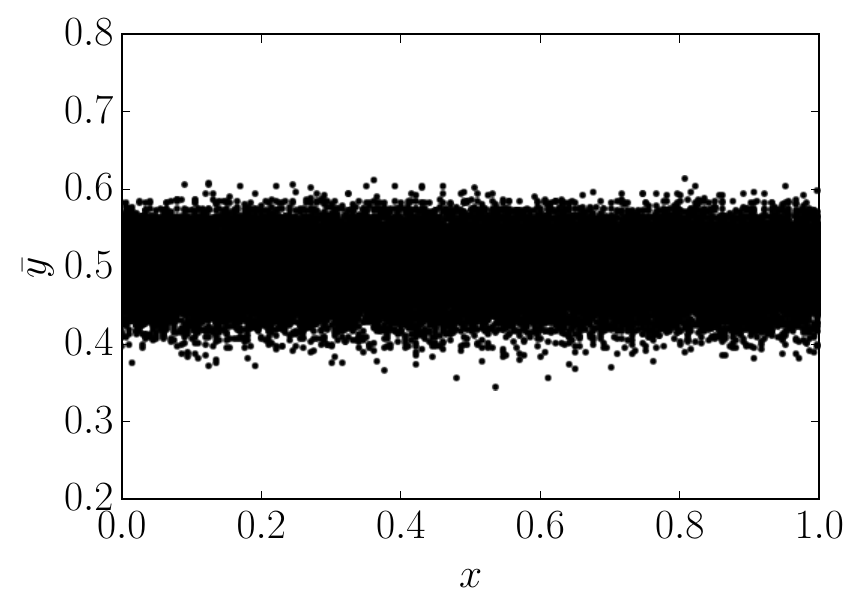}%
    \includegraphics[align=c,width=.19\linewidth]{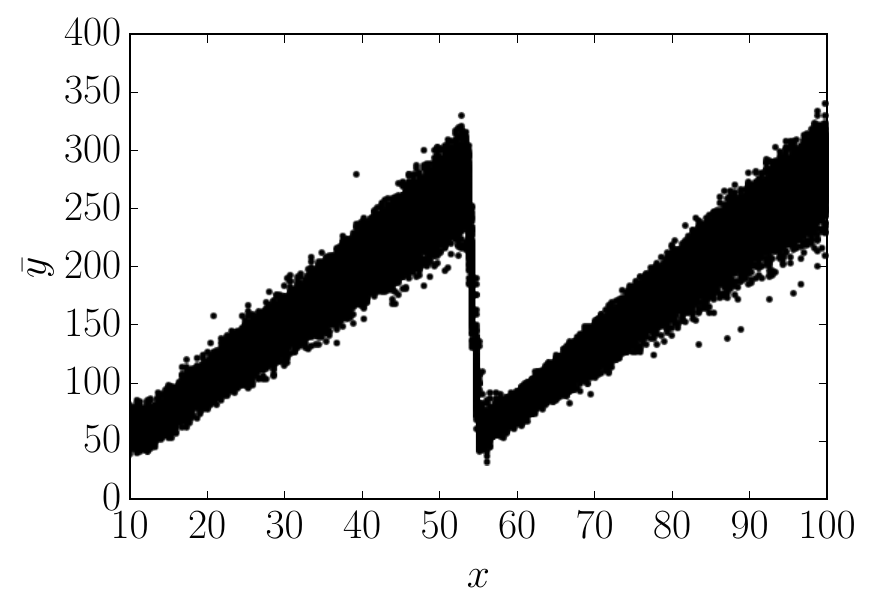}%
    \includegraphics[align=c,width=.19\linewidth]{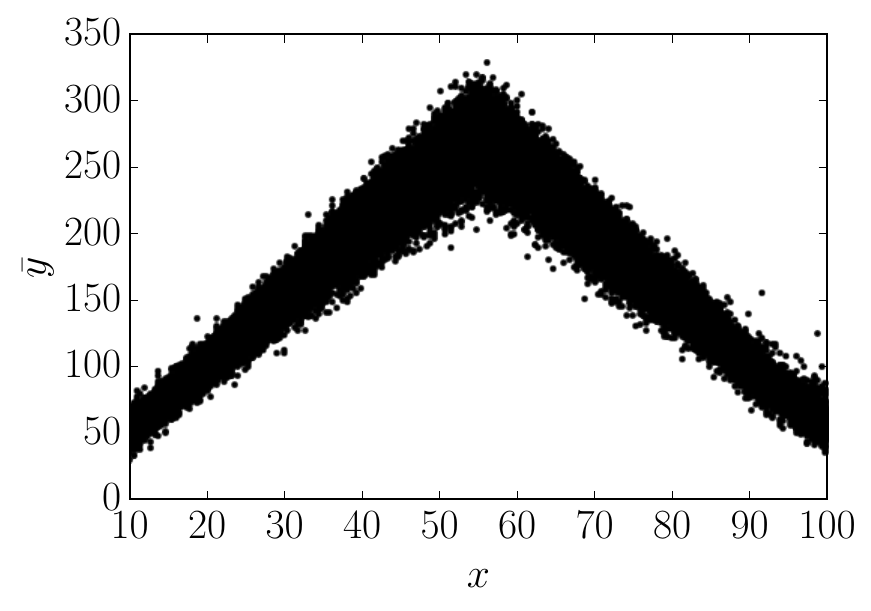}%
    \includegraphics[align=c,width=.19\linewidth]{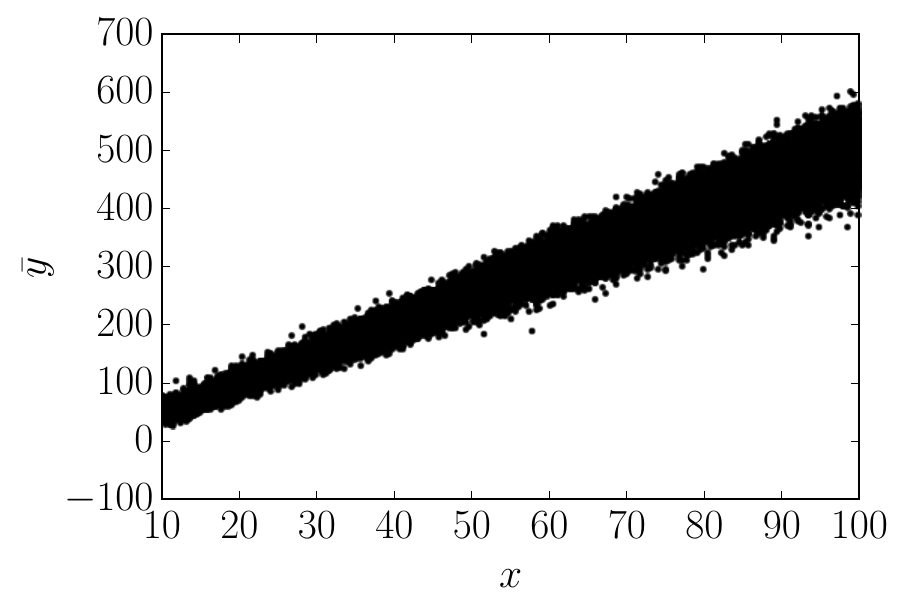}%
    \includegraphics[align=c,width=.19\linewidth]{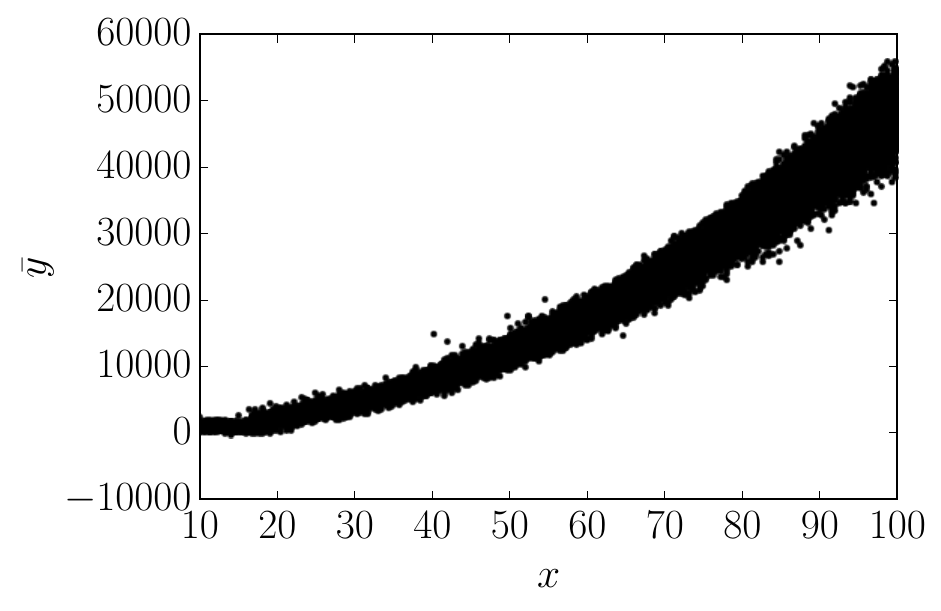}%
    \\
    (e)%
    \includegraphics[align=c,width=.19\linewidth]{plots/diamond_0.2_nobias_3.pdf}%
    \includegraphics[align=c,width=.19\linewidth]{plots/saw_0.2_nobias_3.pdf}%
    \includegraphics[align=c,width=.19\linewidth]{plots/triangle_0.2_nobias_3.pdf}%
    \includegraphics[align=c,width=.19\linewidth]{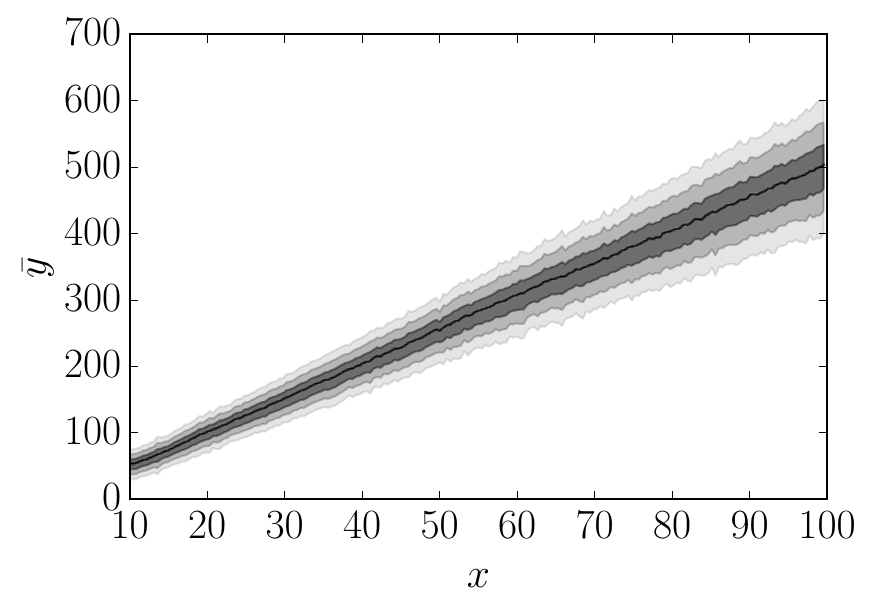}%
    \includegraphics[align=c,width=.19\linewidth]{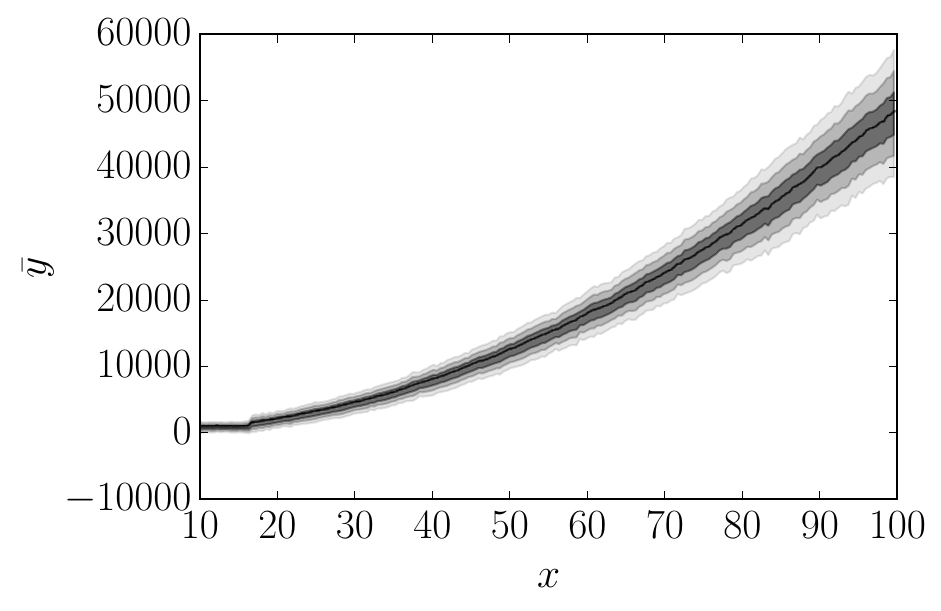}%
    \\
    (f)%
    \includegraphics[align=c,width=.19\linewidth]{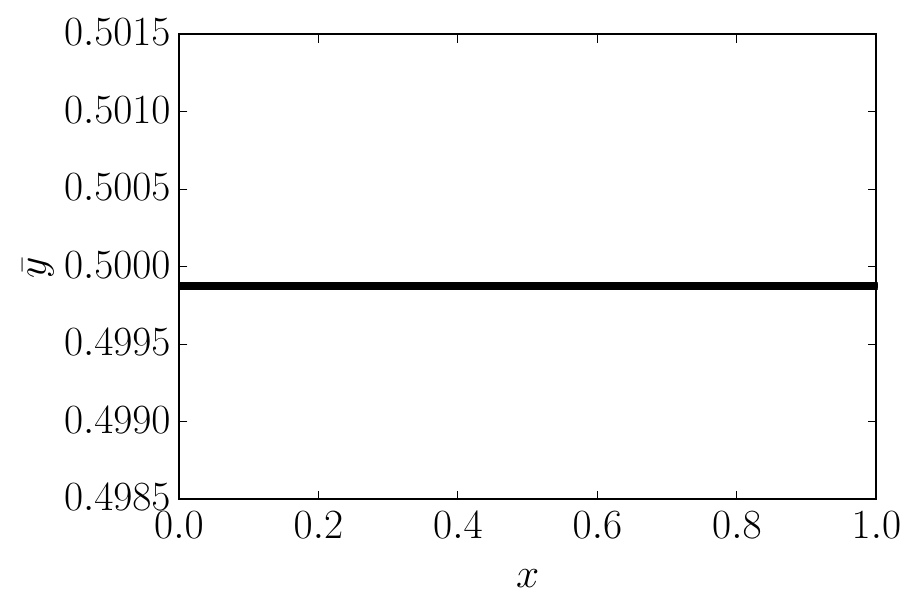}%
    \includegraphics[align=c,width=.19\linewidth]{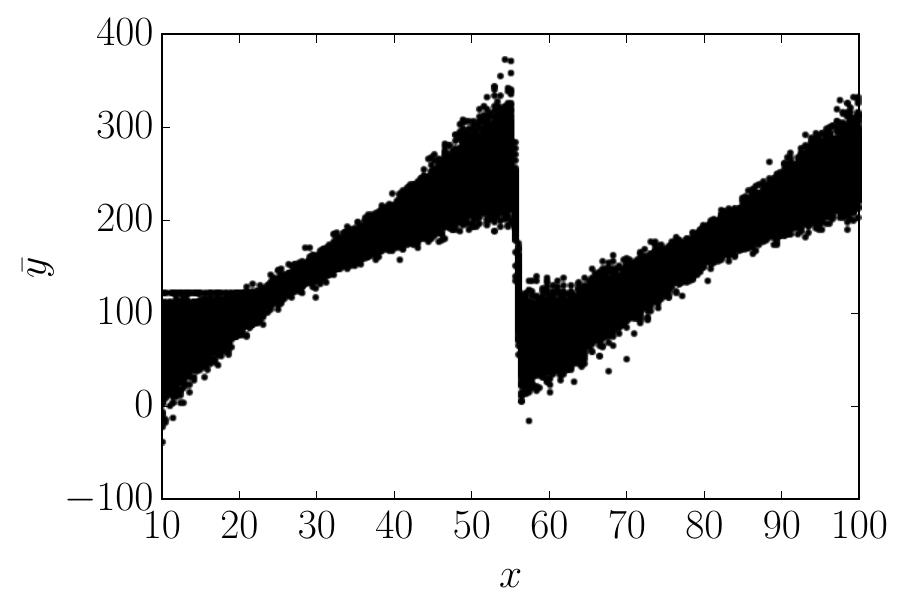}%
    \includegraphics[align=c,width=.19\linewidth]{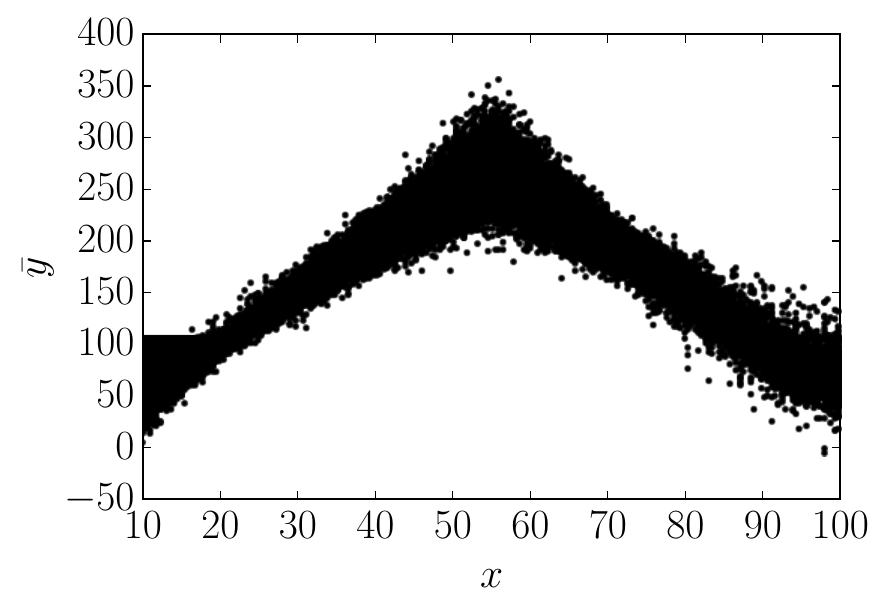}%
    \includegraphics[align=c,width=.19\linewidth]{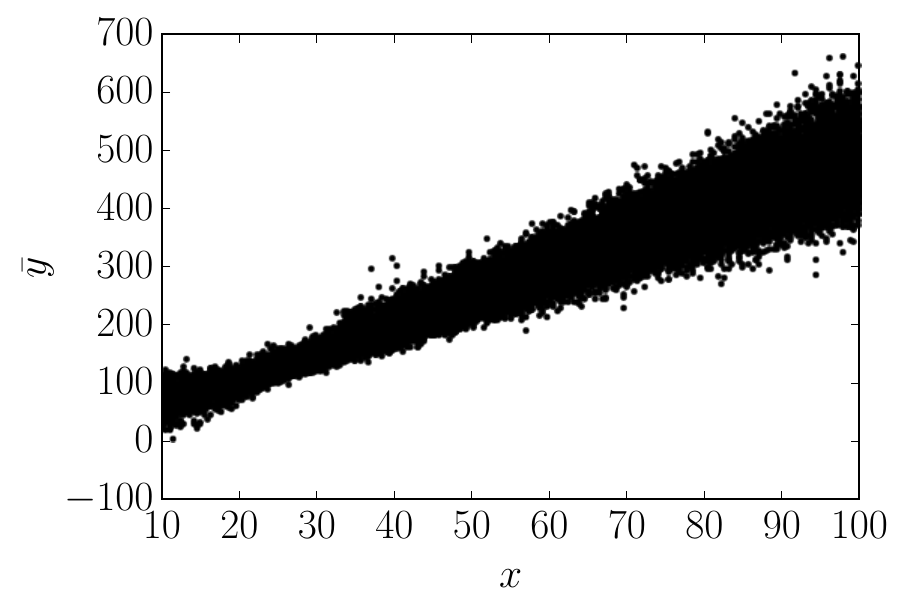}%
    \includegraphics[align=c,width=.19\linewidth]{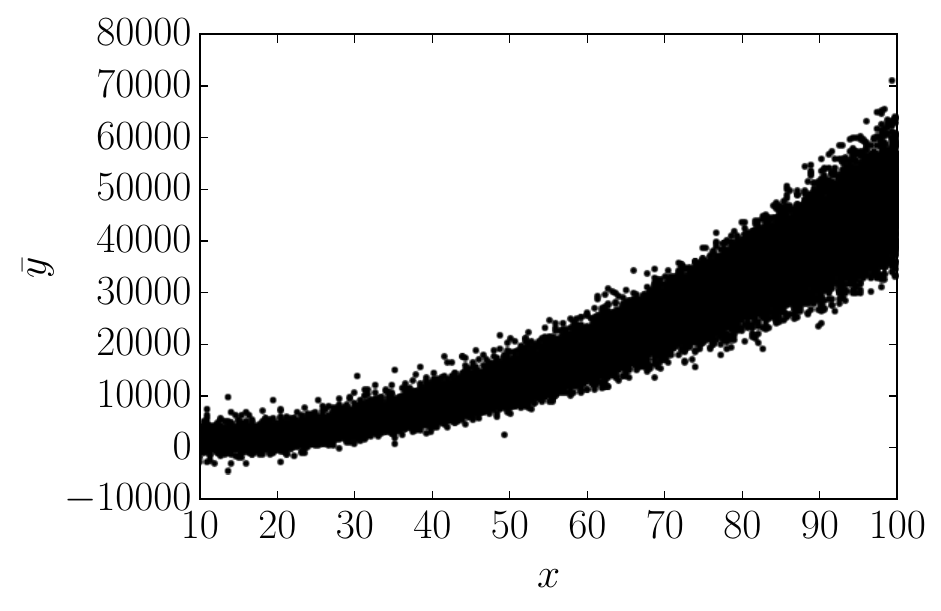}%
    \\
    (g)%
    \includegraphics[align=c,width=.19\linewidth]{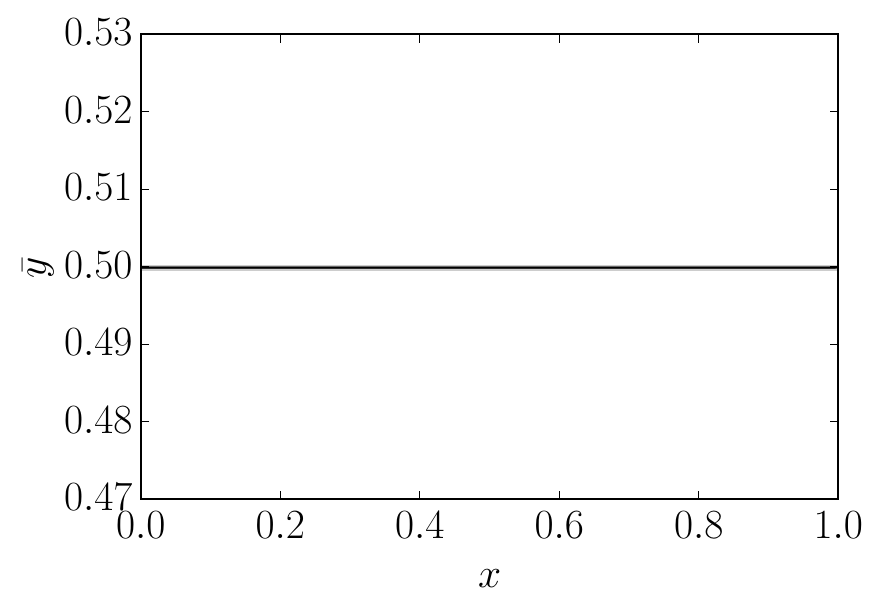}%
    \includegraphics[align=c,width=.19\linewidth]{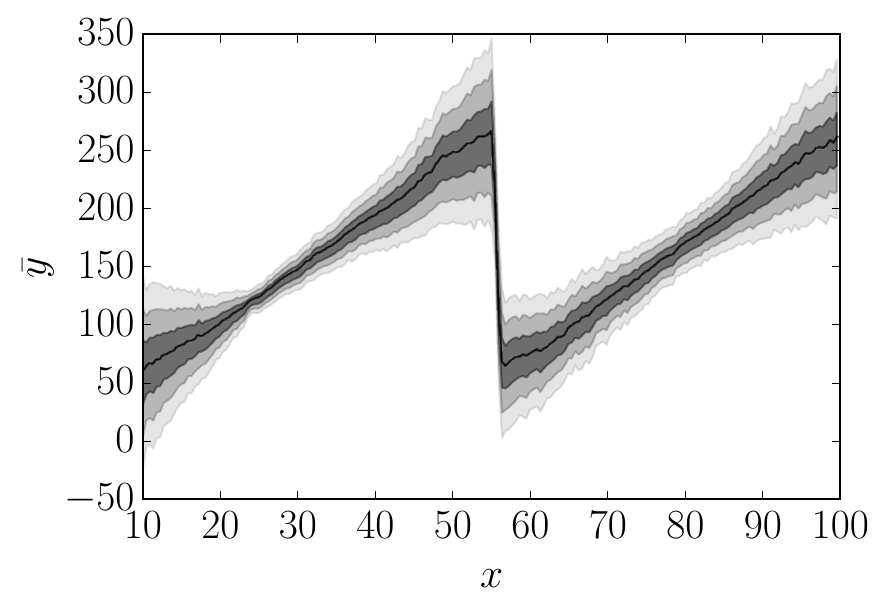}%
    \includegraphics[align=c,width=.19\linewidth]{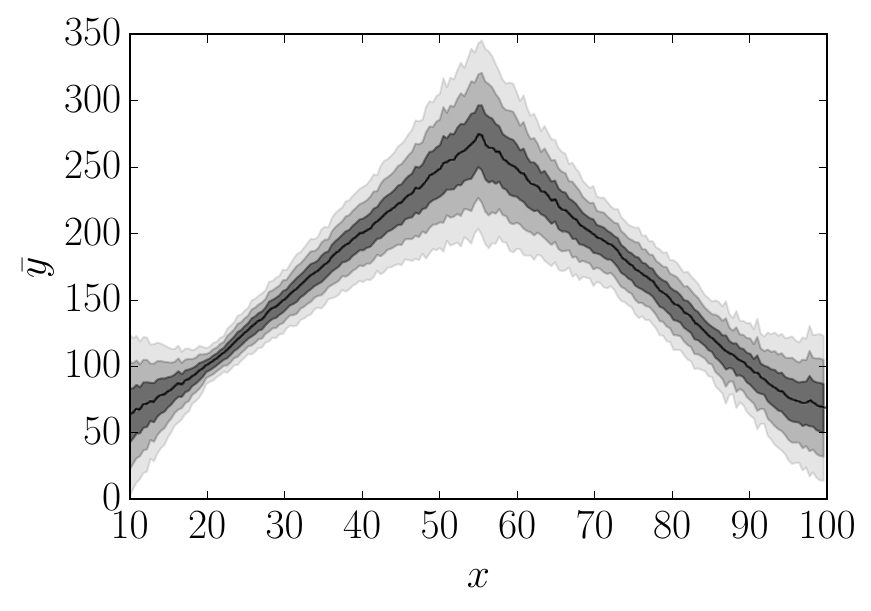}%
    \includegraphics[align=c,width=.19\linewidth]{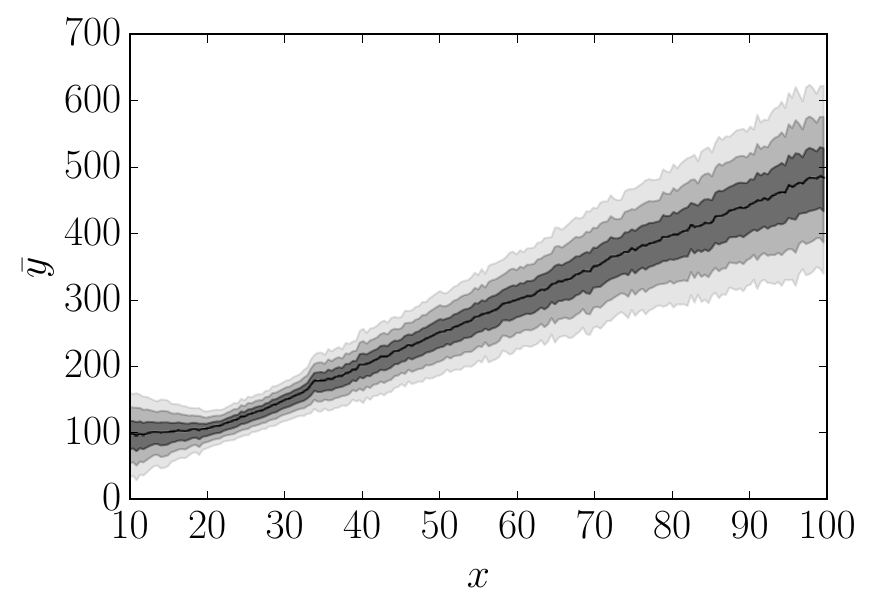}%
    \includegraphics[align=c,width=.19\linewidth]{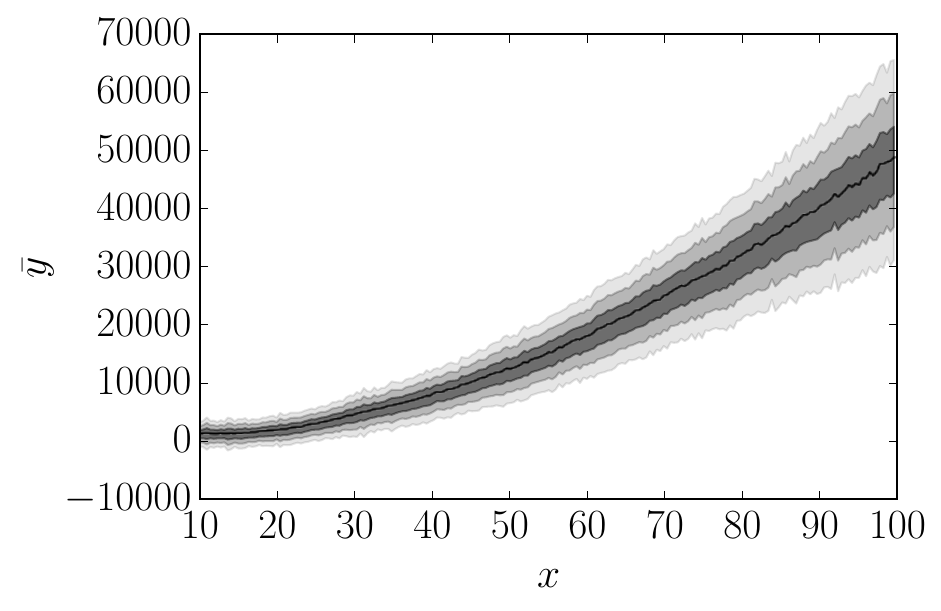}%
    \\
    (h)%
    \includegraphics[align=c,width=.19\linewidth]{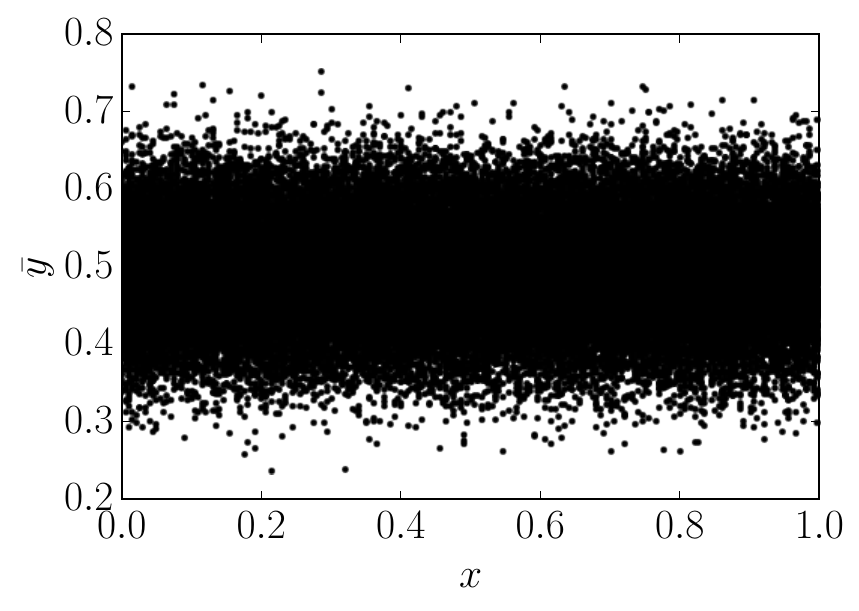}%
    \includegraphics[align=c,width=.19\linewidth]{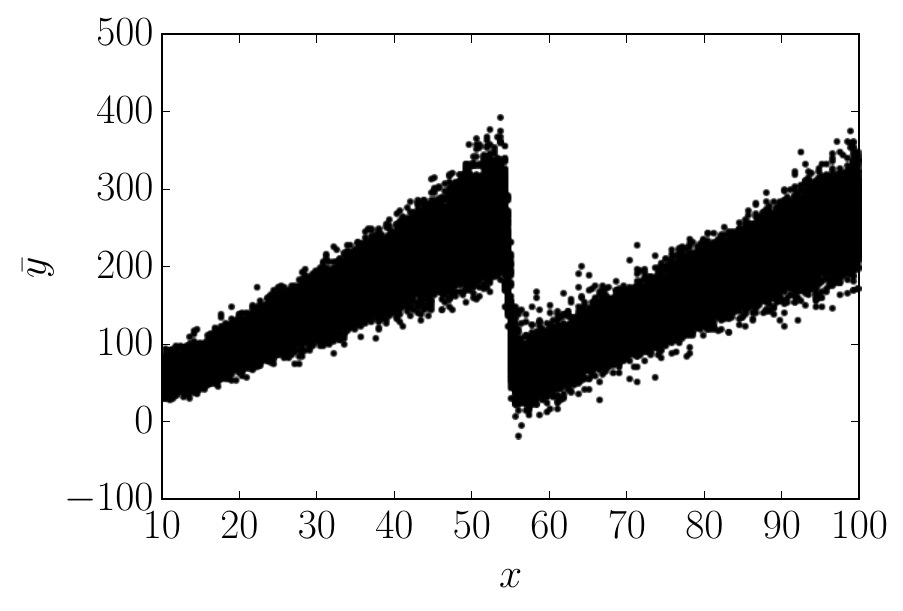}%
    \includegraphics[align=c,width=.19\linewidth]{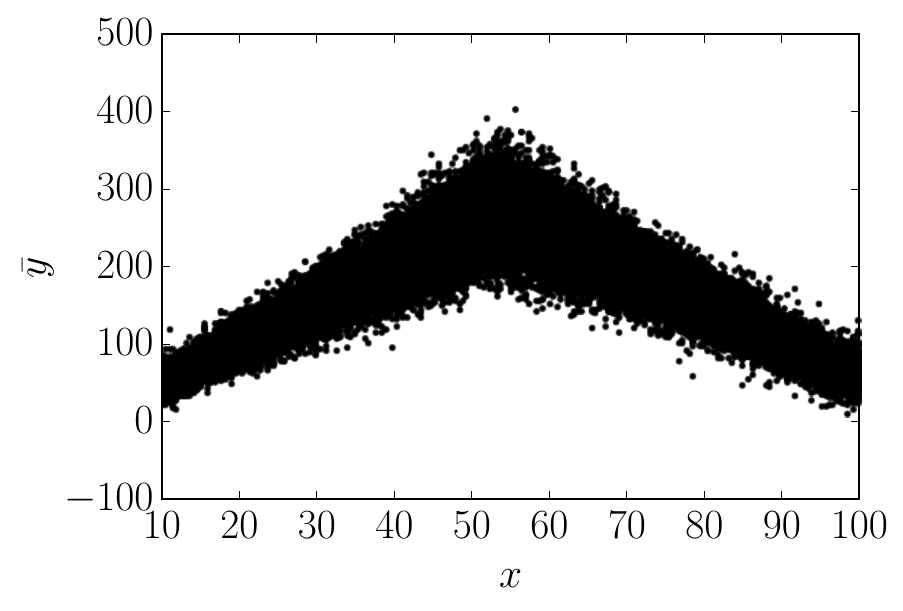}%
    \includegraphics[align=c,width=.19\linewidth]{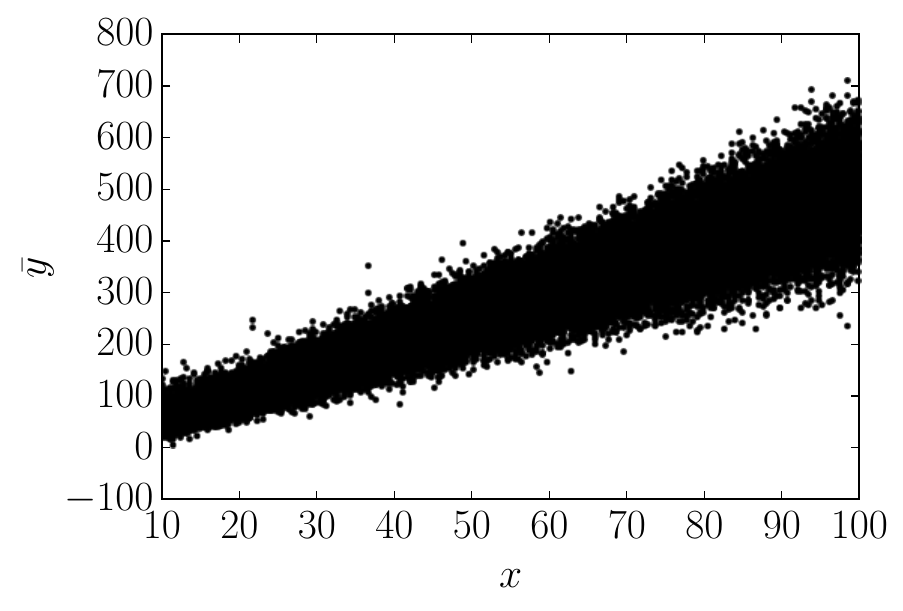}%
    \includegraphics[align=c,width=.19\linewidth]{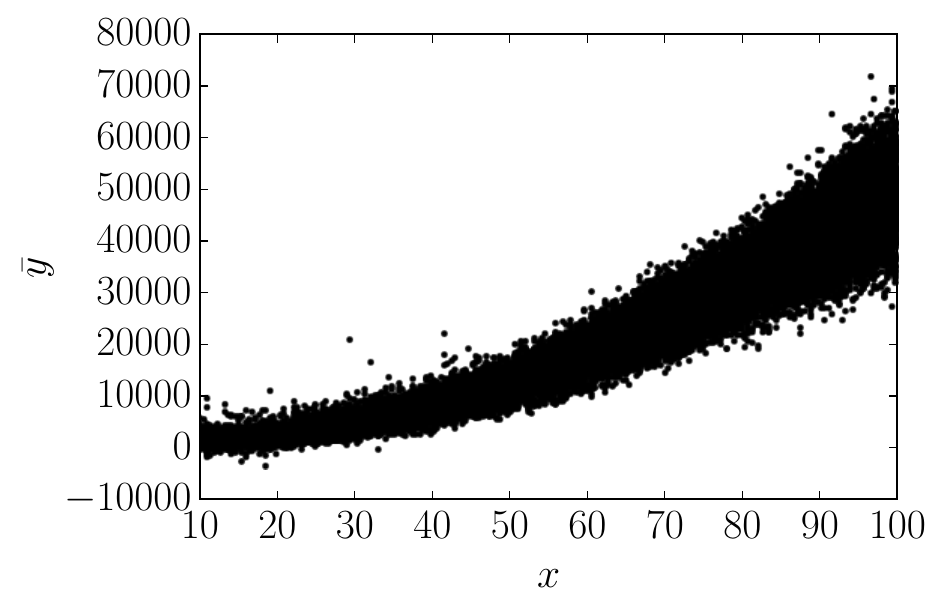}%
    \\
    (i)%
    \includegraphics[align=c,width=.19\linewidth]{plots/diamond_0.5_nobias_3.pdf}%
    \includegraphics[align=c,width=.19\linewidth]{plots/saw_0.5_nobias_3.pdf}%
    \includegraphics[align=c,width=.19\linewidth]{plots/triangle_0.5_nobias_3.pdf}%
    \includegraphics[align=c,width=.19\linewidth]{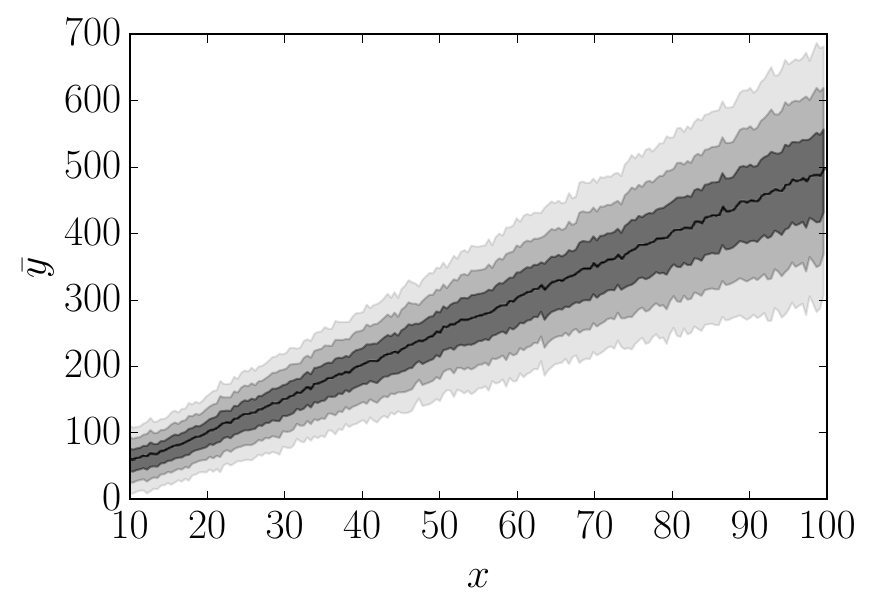}%
    \includegraphics[align=c,width=.19\linewidth]{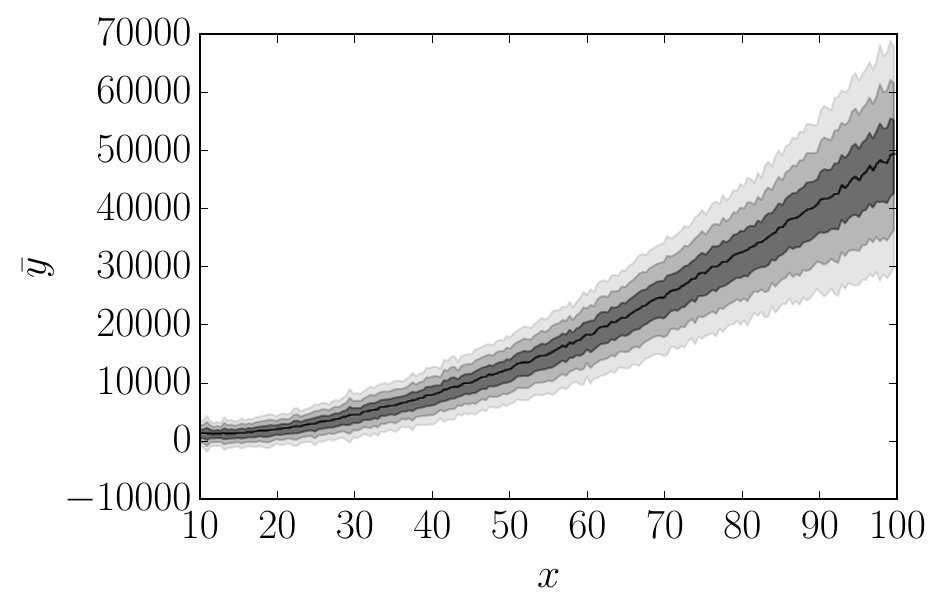}%
\end{figure*}

\end{document}